\documentclass{article}
\usepackage[T1]{fontenc}
\usepackage{microtype}
\usepackage{amsmath,amssymb}
\usepackage{booktabs,array}
\usepackage{multirow}
\usepackage{xcolor}
\usepackage{graphicx}
\usepackage{tikz,pgfplots}
\pgfplotsset{compat=1.18}
\newcommand{\tabstyle}{\footnotesize\setlength{\tabcolsep}{4pt}\renewcommand{\arraystretch}{1.1}\setlength{\abovetopsep}{4pt}}
\usetikzlibrary{arrows.meta,positioning,calc,fit,decorations.pathreplacing,
  shapes.geometric}

\definecolor{paperblue}{HTML}{28649C}
\definecolor{papergray}{HTML}{72777D}
\definecolor{paperlight}{HTML}{F0F2F4}
\definecolor{paperorange}{HTML}{A76320}
\colorlet{uxink}{black!72}          
\colorlet{uxwarntext}{paperorange!75!black} 

\newcommand{\uxsize}{\fontsize{8}{9.6}\selectfont}      
\newcommand{\uxfont}{\uxsize\sffamily}
\newcommand{\uxic}{\fontsize{9}{10.8}\selectfont} 

\newlength{\uxicsize}
\newcommand{\uxicon}[2]{\raisebox{-0.16\uxicsize}{%
  \includegraphics[height=\uxicsize]{figures/icons/#1-#2}}}
\newcommand{\uxdeficon}[2]{\DeclareRobustCommand{#1}[1][ink]{\uxicon{#2}{##1}}}
\uxdeficon{\icUser}{user}
\uxdeficon{\icData}{database}
\uxdeficon{\icLock}{lock}
\uxdeficon{\icChip}{cpu}
\uxdeficon{\icMem}{memory}
\uxdeficon{\icCogs}{gear-six}
\uxdeficon{\icCool}{snowflake}
\uxdeficon{\icTherm}{thermometer}
\uxdeficon{\icBatt}{battery-high}
\uxdeficon{\icScreenOn}{sun}
\uxdeficon{\icScreenOff}{moon}
\uxdeficon{\icSave}{floppy-disk}
\uxdeficon{\icPause}{pause}
\uxdeficon{\icPlay}{play}
\uxdeficon{\icRetry}{arrows-clockwise}   
\uxdeficon{\icWarn}{warning}
\uxdeficon{\icFail}{x-circle}
\uxdeficon{\icSwap}{arrows-left-right}
\uxdeficon{\icFwd}{arrow-right}
\uxdeficon{\icBwd}{arrow-left}
\uxdeficon{\cmark}{check}
\uxdeficon{\xmark}{x}
\uxdeficon{\icRecomp}{arrow-clockwise}   
\uxdeficon{\icGraph}{graph}              
\uxdeficon{\icFuse}{arrows-in}           
\uxdeficon{\icRestore}{clock-counter-clockwise} 
\uxdeficon{\icYield}{user}               
\DeclareRobustCommand{\icNoCool}[1][ink]{\tikz[baseline=(uxnc.base)]{%
  \node[inner sep=0pt,outer sep=0pt](uxnc){\uxicon{snowflake}{#1}};%
  \draw[white,line width=1.9pt,line cap=round]
    ([xshift=1.3pt,yshift=1.3pt]uxnc.south west)--([xshift=-1.3pt,yshift=-1.3pt]uxnc.north east);%
  \draw[uxwarntext,line width=0.9pt,line cap=round]
    ([xshift=1.3pt,yshift=1.3pt]uxnc.south west)--([xshift=-1.3pt,yshift=-1.3pt]uxnc.north east);}}

\tikzset{
  ux/base/.style={font=\uxfont,line width=0.6pt,inner sep=2pt},
  ux/box/.style={draw=papergray,fill=paperlight,line width=0.6pt,
    rounded corners=1.2pt,align=center,inner sep=2pt,font=\uxfont,
    minimum height=0.5cm},
  ux/ours/.style={ux/box,draw=paperblue,fill=paperblue!15,line width=0.8pt},
  ux/warn/.style={ux/box,draw=paperorange,fill=paperorange!15},
  ux/scope/.style={draw=papergray,line width=0.6pt,rounded corners=2pt,
    dash pattern=on 2pt off 1.5pt},
  ux/oursscope/.style={ux/scope,draw=paperblue},
  UxTip/.tip={Latex[length=1.6mm,width=1.4mm]},
  ux/flow/.style={-{UxTip},line width=0.8pt,draw=uxink},
  ux/oursflow/.style={ux/flow,draw=paperblue,line width=1.2pt}, 
  ux/warnflow/.style={ux/flow,draw=paperorange},
  ux/ctrl/.style={-{UxTip},line width=0.5pt,draw=uxink,
    dash pattern=on 2pt off 1.5pt},
  ux/warnctrl/.style={ux/ctrl,draw=paperorange},
  ux/label/.style={font=\uxfont,inner sep=1pt,align=center},
  ux/math/.style={font=\uxsize,inner sep=1pt},
  ux/icon/.style={font=\uxic,inner sep=0.5pt},
  ux/panel/.style={font=\uxfont\bfseries,inner sep=0pt,
    anchor=north west},
  ux/ptitle/.style={font=\uxfont,inner sep=0pt,anchor=base west},
  ux/badge/.style={fill=papergray!22,draw=papergray,line width=0.4pt,
    rounded corners=1.2pt,inner sep=1pt,font=\uxfont\bfseries},
  ux/lane/.style={fill=paperlight,draw=none},
  ux/refmark/.style={diamond,fill=black,draw=white,line width=0.3pt,inner sep=0pt,
    minimum width=4.6pt,minimum height=5.6pt},
  ux/value/.style={font=\uxfont,inner sep=1.5pt},
  ux ours/.style={fill=paperblue!75,draw=paperblue,line width=0.8pt},
  ux oursl/.style={fill=paperblue!22,draw=paperblue,line width=0.5pt},
  ux base0/.style={fill=papergray!20,draw=papergray,line width=0.5pt},
  ux base1/.style={fill=papergray!35,draw=papergray,line width=0.5pt},
  ux base2/.style={fill=papergray!50,draw=papergray,line width=0.5pt},
  ux base3/.style={fill=papergray!65,draw=papergray,line width=0.5pt},
  ux/brace/.style={decorate,decoration={brace,amplitude=2.5pt},
    line width=0.6pt,draw=uxink},
}

\tikzset{
  ux/tile/.code={\pgfqkeys{/ux/tile}{#1}},
  /ux/tile/.cd,
  n/.store in=\uxTileN, n=3,
  cell/.store in=\uxTileCell, cell=0.22cm,
  fill/.store in=\uxTileFill, fill=papergray!30,
  dark/.store in=\uxTileDark, dark=papergray!60,
  mark/.store in=\uxTileMark, mark=none,
  markfill/.store in=\uxTileMarkFill, markfill=paperblue,
}
\tikzset{
  ux/wtile/.pic={
    \begin{scope}[x=\uxTileCell,y=\uxTileCell]
      \pgfmathsetmacro\uxh{\uxTileN/2}
      \pgfmathtruncatemacro\uxnm{\uxTileN-1}
      \fill[\uxTileFill] (-\uxh,-\uxh) rectangle (\uxh,\uxh);
      \foreach \i in {0,...,\uxnm}{\foreach \j in {0,...,\uxnm}{
        \pgfmathtruncatemacro\uxp{mod(\i*7+\j*3+\i*\j,5)}
        \ifnum\uxp<2\relax
          \fill[\uxTileDark] (-\uxh+\i,-\uxh+\j) rectangle ++(1,1);
        \fi}}
      \def\uxrow{row}\def\uxcol{col}
      \ifx\uxTileMark\uxrow
        \fill[\uxTileMarkFill] (-\uxh,\uxh-1) rectangle (\uxh,\uxh);\fi
      \ifx\uxTileMark\uxcol
        \fill[\uxTileMarkFill] (-\uxh,-\uxh) rectangle (-\uxh+1,\uxh);\fi
      \ifnum\uxnm>0\relax
        \foreach \k in {1,...,\uxnm}{
          \draw[white,line width=0.35pt] (-\uxh+\k,-\uxh) -- (-\uxh+\k,\uxh);
          \draw[white,line width=0.35pt] (-\uxh,-\uxh+\k) -- (\uxh,-\uxh+\k);}
      \fi
      \draw[papergray!80!black,line width=0.5pt] (-\uxh,-\uxh) rectangle (\uxh,\uxh);
      \path (-\uxh,-\uxh) coordinate (-sw) (\uxh,\uxh) coordinate (-ne);
    \end{scope}
    \node[inner sep=0pt,outer sep=0pt,fit=(-sw)(-ne)] (-box) {};
  },
  ux/qtile/.pic={
    \begin{scope}[x=\uxTileCell,y=\uxTileCell]
      \pgfmathsetmacro\uxh{\uxTileN/2}
      \pgfmathtruncatemacro\uxnm{\uxTileN-1}
      \fill[\uxTileFill] (-\uxh,-\uxh) rectangle (\uxh,\uxh);
      \foreach \i in {0,...,\uxnm}{\foreach \j in {0,...,\uxnm}{
        \pgfmathtruncatemacro\uxp{mod(\i*7+\j*3+\i*\j,5)}
        \ifnum\uxp<2\relax
          \fill[\uxTileDark] (-\uxh+\i,-\uxh+\j) rectangle ++(1,1);
        \fi}}
      \fill[papergray!85!black] (\uxh+0.35,-\uxh) rectangle (\uxh+1.15,\uxh);
      \foreach \k in {1,...,\uxnm}{
        \draw[white,line width=0.3pt] (-\uxh+\k,-\uxh) -- (-\uxh+\k,\uxh);
        \draw[white,line width=0.3pt] (-\uxh,-\uxh+\k) -- (\uxh+1.15,-\uxh+\k);}
      \draw[papergray!80!black,line width=0.5pt] (-\uxh,-\uxh) rectangle (\uxh,\uxh);
      \draw[papergray!80!black,line width=0.5pt] (\uxh+0.35,-\uxh) rectangle (\uxh+1.15,\uxh);
      \path (-\uxh,-\uxh) coordinate (-sw) (\uxh+1.15,\uxh) coordinate (-ne);
    \end{scope}
    \node[inner sep=0pt,outer sep=0pt,fit=(-sw)(-ne)] (-box) {};
  },
}
\tikzset{
  ux/lora/.pic={
    \fill[paperblue!80] (-0.15,-0.19) -- (0.15,-0.19) -- (0.04,-0.02)
      -- (-0.04,-0.02) -- cycle;
    \fill[paperblue!80] (-0.04,0.02) -- (0.04,0.02) -- (0.15,0.19)
      -- (-0.15,0.19) -- cycle;
    \path (-0.15,-0.19) coordinate (-sw) (0.15,0.19) coordinate (-ne);
    \node[inner sep=0pt,outer sep=0pt,fit=(-sw)(-ne)] (-box) {};
  },
}
\tikzset{
  ux/rows/.code={\pgfqkeys{/ux/rows}{#1}},
  /ux/rows/.cd,
  list/.store in=\uxRowList, list={0.5,0.3,0.6},
  fill/.store in=\uxRowFill, fill=paperblue!55,
  h/.store in=\uxRowH, h=0.075cm,
  gap/.store in=\uxRowGap, gap=0.045cm,
  hl/.store in=\uxRowHl, hl=0,
  hlfill/.store in=\uxRowHlFill, hlfill=paperorange,
}
\tikzset{
  ux/rows/.pic={
    \foreach \l [count=\i] in \uxRowList {
      \ifnum\i=\uxRowHl\relax\def\uxrf{\uxRowHlFill}\else\def\uxrf{\uxRowFill}\fi
      \fill[\uxrf] (0,{-(\i-1)*(\uxRowH+\uxRowGap)}) rectangle
        (\l,{-(\i-1)*(\uxRowH+\uxRowGap)-\uxRowH});
      \xdef\uxRowCount{\i}
    }
    \path (0,0) coordinate (-nw)
      (0,{-\uxRowCount*(\uxRowH+\uxRowGap)+\uxRowGap}) coordinate (-sw);
  },
}

\newcommand{\uxbarw}{3.5pt}
\newcommand{\uxhbar}[4]{\draw[#4] ([yshift=-\uxbarw]axis cs:#2,#1) rectangle
  ([yshift=\uxbarw]axis cs:#3,#1);}
\newcommand{\uxvbar}[4]{\draw[#4] ([xshift=#2-\uxbarw]axis cs:#1,0) rectangle
  ([xshift=#2+\uxbarw]axis cs:#1,#3);}

\newcommand{\uxdiamond}{\raisebox{0.2pt}{$\blacklozenge$}}

\pgfplotsset{
  ux axis/.style={
    scale only axis,font=\uxfont,
    xticklabel={\pgfmathprintnumber[assume math mode=true]{\tick}},
    yticklabel={\pgfmathprintnumber[assume math mode=true]{\tick}},
    axis line style={draw=papergray,line width=0.5pt},
    tick style={draw=papergray,line width=0.5pt},tick align=outside,
    major tick length=2pt,
    axis x line*=bottom,axis y line*=left,
    grid style={draw=black!12,line width=0.3pt},
    label style={font=\uxfont},tick label style={font=\uxfont},
    title style={font=\uxfont},
    every axis label/.append style={inner sep=1pt},
    xlabel style={yshift=2pt},ylabel style={yshift=-2pt},
    clip=false,
    legend style={font=\uxfont,draw=none,fill=none,inner sep=0pt,
      /tikz/every even column/.append style={column sep=5pt}},
    legend cell align=left,
    legend image post style={line width=0.6pt},
  },
  ux hbar/.style={ux axis,xbar,bar width=7pt,bar shift=0pt,
    xmajorgrids=true,y axis line style={draw=none},ytick style={draw=none},
    ytick=data,enlarge y limits={abs=0.45}},
  ux vbar/.style={ux axis,ybar,bar width=7pt,
    ymajorgrids=true,xtick style={draw=none}},
  ux ref/.style={only marks,mark=diamond*,mark size=2pt,
    mark options={fill=black,draw=white,line width=0.3pt}},
}

\DeclareMathSizes{8}{8}{7}{5}

\usepackage{balance}

\usepackage[onecolumn]{trustai4s}  
\usepackage{hyperref}
\labpreprintbrand{}{SMU-FIU TrustAI4S Lab}{}
\usepackage{xurl}
\usepackage{enumitem}
\mlsystitlerunning{First-Order LoRA Training with Phone NPUs}
\makeatletter
\renewcommand{\Notice@String}{Preprint.}
\makeatother
\hypersetup{
  pdftitle={First-Order LoRA Training with Phone NPUs},
  pdfauthor={Xu Zheng, Wei Cheng, Zhuomin Chen, Dongkuan Xu, Mo Sha, Jingchao Ni, Dongsheng Luo}}
\newcommand{\dx}{\mathrm{d}X}
\newcommand{\dy}{\mathrm{d}Y}
\begin{document}
\twocolumn[
\mlsystitle{First-Order LoRA Training with Phone NPUs}
\begin{mlsysauthorlist}
\mlsysauthor{Xu Zheng}{fiu}
\mlsysauthor{Wei Cheng}{nec}
\mlsysauthor{Zhuomin Chen}{fiu}
\mlsysauthor{Dongkuan Xu}{ncsu}
\mlsysauthor{Mo Sha}{fiu}
\mlsysauthor{Jingchao Ni}{uh}
\mlsysauthor{Dongsheng Luo}{smu}
\end{mlsysauthorlist}
\mlsysaffiliation{fiu}{Florida International University, Miami, FL, USA}
\mlsysaffiliation{nec}{NEC Laboratories America, Princeton, NJ, USA}
\mlsysaffiliation{ncsu}{North Carolina State University, Raleigh, NC, USA}
\mlsysaffiliation{uh}{University of Houston, Houston, TX, USA}
\mlsysaffiliation{smu}{Singapore Management University, Singapore}
\mlsyscorrespondingauthor{Dongsheng Luo}{luodongsheng01@gmail.com}
\mlsyskeywords{on-device fine-tuning, large language models, LoRA, mobile NPU, mixed precision}
\vskip 0.3in
\begin{abstract}

On-device fine-tuning lets a large language model (LLM) learn from its user's private data without uploading it.
Low-rank adaptation (LoRA) reduces the trainable parameters, but each first-order training step still multiplies by the frozen base-model weights during the forward pass, activation recomputation, and input-gradient computation.
A phone's neural processing unit (NPU) already holds these weights for inference; using it for training also requires computing input gradients, which its inference interface does not support.
We present a system that runs all three frozen-weight products on the NPU while keeping the training graph, adapters, and optimizer on the CPU.
Its input-gradient kernels reuse the resident weight tiles and the forward dequantizer, avoiding a transposed weight copy and ensuring forward and backward use the same effective weights.
To preserve usable gradients, the system selects precision by gradient sensitivity and applies a per-row range guard for half-precision (FP16) outputs.
Its runtime fuses the remaining CPU operators and overlaps NPU execution with adapter computation, while resumable execution and thermal pacing let training proceed as a background job.
We evaluate it on two Qualcomm Snapdragon platforms using four LLMs from the Qwen and Llama families.
Our workloads cover nightly personalization with labels supplied on the phone and the public LaMP-2 benchmark; comparisons include an optimized CPU trainer, a phone-GPU trainer, forward-only training, and GPU-trained references.
\end{abstract}
]
\printAffiliationsAndNotice{}

\section{Introduction}\label{sec:intro}

Phones now run large language models (LLMs) locally, and deployed assistants specialize a shared base model with low-rank adaptation (LoRA)~\cite{applefm}.
Such a model serves its user better once it learns from that user's private examples, which accumulate on the phone.
Training where the data lives avoids uploading it and leaves room for federated aggregation~\cite{federated}.
Making on-device adaptation practical requires training to fit within the phone's compute, memory, and thermal limits.

\begin{figure*}[t]
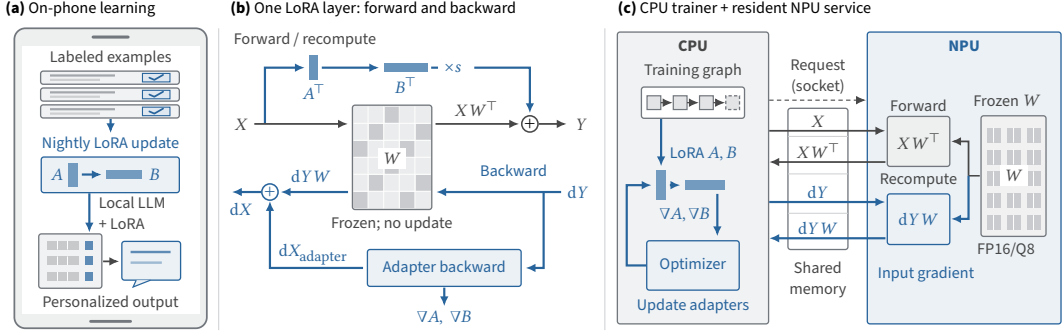

\centering
\begingroup
\tikzset{
  ov2/text/.style={font=\uxfont,text=uxink,inner sep=1pt,align=center},
  ov2/math/.style={font=\uxsize,text=uxink,inner sep=1pt},
  ov2/box/.style={draw=papergray,fill=white,line width=.6pt,rounded corners=1.2pt},
  ov2/bluebox/.style={ov2/box,draw=paperblue,fill=paperblue!9},
  ov2/base/.style={ux/flow,line width=.8pt},
  ov2/learn/.style={ux/oursflow,line width=1pt},
  ov2/sum/.style={circle,draw=uxink,fill=white,line width=.7pt,
    minimum size=.26cm,inner sep=0pt,font=\uxsize},
}
\resizebox{\textwidth}{!}{%
\begin{tikzpicture}[x=1cm,y=1cm]
\path[use as bounding box] (0,0) rectangle (16.9,5.4);
\node[ux/ptitle] at (0,5.12) {\textbf{(a)} On-phone learning};
\node[ux/ptitle] at (3.60,5.12) {\textbf{(b)} One LoRA layer: forward and backward};
\node[ux/ptitle] at (9.80,5.12) {\textbf{(c)} CPU trainer + resident NPU service};
\draw[papergray!35,line width=.4pt] (3.42,.05) -- (3.42,4.88);
\draw[papergray!35,line width=.4pt] (9.60,.05) -- (9.60,4.88);
\begin{scope}
\draw[ov2/box,fill=papergray!12,rounded corners=5pt,line width=.8pt]
  (.15,.08) rectangle (3.23,4.88);
\fill[white] (.27,.31) rectangle (3.11,4.59);
\draw[papergray,line cap=round,line width=1.4pt] (1.41,4.74)--(1.97,4.74);
\draw[papergray,line cap=round,line width=1.3pt] (1.37,.19)--(2.01,.19);
\node[ov2/text] at (1.69,4.41) {Labeled examples};
\foreach \y in {3.55,3.82,4.09}{
  \draw[ov2/box,fill=paperlight] (.57,\y-.105) rectangle (2.79,\y+.105);
  \draw[papergray!65,line width=1pt] (.72,\y+.035)--(1.76,\y+.035);
  \draw[papergray!35,line width=1pt] (.72,\y-.035)--(1.52,\y-.035);
  \draw[paperblue,fill=paperblue!15,line width=.5pt]
    (2.20,\y-.068) rectangle (2.64,\y+.068);
  \draw[paperblue,line width=.8pt,line cap=round,line join=round]
    (2.31,\y)--(2.39,\y-.035)--(2.53,\y+.045);
}
\draw[ov2/learn] (1.69,3.43)--(1.69,3.20);
\node[ov2/text,text=paperblue] at (1.69,3.02) {Nightly LoRA update};
\draw[ov2/bluebox] (.58,2.25) rectangle (2.79,2.82);
\fill[paperblue!65] (1.01,2.35) rectangle (1.16,2.72);
\fill[paperblue!65] (1.62,2.47) rectangle (2.19,2.60);
\node[ov2/math,text=paperblue] at (.83,2.54) {$A$};
\node[ov2/math,text=paperblue] at (2.39,2.54) {$B$};
\draw[ov2/learn] (1.23,2.54)--(1.55,2.54);
\draw[ov2/learn] (1.35,2.24)--(1.35,1.61);
\node[ov2/text,anchor=west,align=left] at (1.46,1.92) {Local LLM\\+ LoRA};
\draw[ov2/box,fill=paperlight] (.54,.72) rectangle (1.55,1.59);
\foreach \y in {.84,1.09,1.34}{
  \fill[papergray!40] (.68,\y) rectangle (1.18,\y+.14);
  \draw[white,line width=.35pt] (.84,\y)--(.84,\y+.14) (1.00,\y)--(1.00,\y+.14);
  \fill[paperblue!75] (1.29,\y) rectangle (1.41,\y+.14);
}
\draw[ov2/base] (1.57,1.15)--(1.83,1.15);
\draw[ov2/bluebox] (1.87,.85) rectangle (2.81,1.47);
\draw[paperblue,line width=.55pt,fill=paperblue!9]
  (2.10,.85)--(2.05,.74)--(2.33,.85);
\draw[paperblue!75,line width=1.1pt] (2.01,1.30)--(2.65,1.30);
\draw[paperblue!45,line width=1.1pt] (2.01,1.13)--(2.51,1.13);
\node[ov2/text] at (1.69,.49) {Personalized output};
\end{scope}
\begin{scope}[shift={(3.60,0)}]
  \input{figures/redraw/overview-layer}
\end{scope}
\begin{scope}[shift={(9.80,0)}]
  \input{figures/redraw/overview-partition}
\end{scope}
\end{tikzpicture}}%
\endgroup
\vspace{-1em}
\caption{Our system fine-tunes the on-device LLM on the user's private data. (a) A nightly update trains its LoRA adapter; (b) the same frozen $W$ serves forward, recomputation, and backward, and the CPU adds $\dy W$ to the adapter input gradient; (c) the CPU trainer and the NPU service run as separate processes on the phone: the CPU owns the graph and optimizer, the NPU holds one weight copy, messages over a local Unix-domain socket carry projection requests, and shared memory carries their input and output tensors.}
\label{fig:overview}
\vspace{-1em}
\end{figure*}

LoRA reduces training memory by freezing the base model and updating only small adapters~\cite{lora}. However, the frozen layers still do much of the computation. Each frozen projection uses its weights $W$ three times per training step: once in the forward pass, once to recompute activations during checkpointing~\cite{checkpoint}, and once to compute the input gradient $\dy W$ during backpropagation. We call these three matrix multiplications the frozen path; all three read the same weights (Figure~\ref{fig:overview}b). Even with LoRA, each training step still requires all three frozen-path multiplications (Section~\ref{sec:frozenpath}).

The phone's neural processing unit (NPU) is built for these matrix products and already serves on-device LLM inference~\cite{llmnpu}. Its inference engine computes the first two products from resident weight tiles, and the third is the same product with the operand roles exchanged. The software stack, however, exposes only inference: forward products with resident, quantized weights and half-precision (FP16) outputs. Existing routes give up backpropagation, the NPU, or model scale. Forward-only methods run on inference engines because they estimate gradients from perturbed forward passes~\cite{fwdllm,mobizo,mobiedit,mezo}. First-order LLM fine-tuning has run on an embedded GPU~\cite{pockengine} and phone GPUs~\cite{qvac}, leaving the NPU idle. Mandheling~\cite{mandheling} and HexaTrain~\cite{hexatrain} train on mobile digital signal processors (DSPs) and NPUs, but only networks far smaller than billion-parameter LLMs. If the NPU also computes $\dy W$ from its resident tiles, the frozen path can run there, while the small adapter and its optimizer state stay on the CPU (Figure~\ref{fig:overview}c).

Building a trainer around this idea presents three challenges.
\emph{Direction:} inference kernels and weight layouts serve $XW^\top$, whereas backward traverses the other dimension; a separately quantized transpose would add resident storage and change the effective weights the gradient sees. 
\emph{Precision and range:} in the engine's FP16 results, small gradients underflow and partial sums can overflow, and inference weight formats target outputs, not gradients.
In models that interleave recurrent Gated DeltaNet (GDN) layers with attention, such as Qwen3.5~\cite{qwen35,gateddelta}, quantization error in a recurrent projection enters a state that every later token reuses. 
\emph{System cost:} once the NPU runs the frozen path, the remaining CPU operators and the CPU--NPU boundary limit each step, and an update runs for hours on a phone whose memory is shared, whose temperature rises, and which its user continues to use.

We present a system that trains an on-device LLM with first-order LoRA and runs the frozen path on the NPU's inference engine. Its input-gradient kernels read resident FP16 tiles and 8-bit block-quantized (Q8) tiles, which store each block of 32 weights as 8-bit integers with one FP16 scale; for Q8, they exchange the matrix engine's operand roles so the forward dequantizer yields $W^\top$. To keep gradients usable, a gradient-sensitive policy keeps recurrent projections in FP16 and activations in floating point, and a range guard scales each gradient row by a power of two, with overflow flags and bounded retries. On the CPU side, fused operators shrink residual work, asynchronous requests overlap NPU products with adapter gradients, and the job commits at optimizer-step boundaries to yield to the user, pause under thermal limits, and resume. Our contributions are:
\begin{itemize}[itemsep=1.5pt,topsep=0pt,parsep=0pt,leftmargin=*]
\item A split of first-order LoRA training between a CPU training graph and a resident NPU service that runs the frozen path from the inference weight tiles (Sections~\ref{sec:overview} and~\ref{sec:impl}).
\item Frozen-path backpropagation on an inference NPU: input-gradient kernels that reuse the resident weights and the forward dequantizer, a gradient-sensitive precision policy, and a per-row range guard (Section~\ref{sec:backward}).
\item A CPU--NPU training runtime for hours-long background jobs: fused residual operators, asynchronous NPU requests, a checkpoint budget, and step-boundary commit, yield, and thermal pacing (Section~\ref{sec:runtime}).
\item An evaluation on two Qualcomm Snapdragon platforms (8 Elite and 8 Elite Gen 5, with Hexagon V79 and V81 NPUs) with four LLMs from two model families, on a nightly personalization update with on-phone labels and on LaMP-2~\cite{lamp}, against an optimized CPU trainer, a phone-GPU trainer, forward-only training, and GPU-trained references (Section~\ref{sec:eval}).
\end{itemize}

\section{Background and Motivation}\label{sec:motivation}

\subsection{Frozen-Weight Computation}\label{sec:frozenpath}

For a frozen projection $W\in\mathbb{R}^{n_{\rm out}\times n_{\rm in}}$ and tokens $X\in\mathbb{R}^{M\times n_{\rm in}}$, a LoRA layer~\cite{lora} computes
\begin{equation}\label{eq:lora}
Y=XW^\top+s(XA^\top)B^\top,
\end{equation}
with adapters $A\in\mathbb{R}^{r\times n_{\rm in}}$ and $B\in\mathbb{R}^{n_{\rm out}\times r}$ and scale $s$. Its base input gradient is
\begin{equation}\label{eq:dx}
\dx_{\rm base}=\dy W.
\end{equation}
Freezing $W$ removes its parameter gradient and optimizer state, but not its products: $XW^\top$ in forward, again when checkpointing recomputes the layer~\cite{checkpoint}, and $\dy W$ for the input gradient. Each costs $n_{\rm in}n_{\rm out}$ multiply-adds per token, against $r(n_{\rm in}+n_{\rm out})$ for the adapter branch, with $r$ far below both (Section~\ref{sec:setup}). Because the frozen path dominates the arithmetic, a projection-level interface to the NPU suffices; the training graph can stay on the CPU.

\subsection{On-Device Training Bottleneck}

\begin{figure}[t]
\centering
\tikzset{
  chart base/.style={fill=papergray!20,draw=papergray,line width=0.5pt},
  chart base2/.style={fill=papergray!50,draw=papergray,line width=0.5pt},
  chart ours/.style={fill=paperblue,draw=paperblue,line width=0.5pt},
  chart warn/.style={fill=paperorange!50,draw=paperorange,line width=0.5pt},
  chart band/.style={fill=paperorange!15,draw=none},
  chart line ours/.style={draw=paperblue,line width=1pt},
  chart line base/.style={draw=papergray,line width=1pt,dash pattern=on 2.5pt off 1.5pt},
  chart line ref/.style={draw=uxink,line width=0.6pt,dash pattern=on 2pt off 1.5pt,
    preaction={draw=white,line width=1pt,solid}},
  chart val/.style={ux/value,anchor=west},
  chart val ours/.style={ux/value,anchor=west,text=paperblue,font=\uxfont\bfseries},
  chart title/.style={font=\uxfont,inner sep=0pt,anchor=base},
}
\newcommand{\ftitledrop}{20pt}
\newcommand{\ftitle}[3]{\node[chart title] at ([yshift=-\ftitledrop]#1.south)
  {\textbf{(#2)}\hspace{3pt}#3};}
\newcommand{\chartkey}[2]{{\setlength{\fboxsep}{0pt}\setlength{\fboxrule}{0.5pt}%
  \fcolorbox{#2}{#1}{\rule{0pt}{4.5pt}\rule{4.5pt}{0pt}}}}
\newcommand{\fgrid}[3]{\draw[/pgfplots/every axis grid]
  ({axis cs:#1,0} |- #2) -- ({axis cs:#1,0} |- #3);}
\colorlet{motwarn}{paperblue!60}
\colorlet{motwarntext}{paperblue!75}
\tikzset{chart warn/.style={fill=paperblue!40,draw=motwarn,line width=0.5pt}}
\newcommand{\motW}{7.3cm}
\newcommand{\mg}[3]{\fgrid{#1}{{axis cs:0,#2}}{{axis cs:0,#3}}}
\newcommand{\mlab}[5][]{\node[ux/label,inner xsep=0pt,inner ysep=1.5pt,anchor=base,#1]
  (#5) at (axis cs:#2,#3) {#4};}
\newcommand{\motPanelA}{%
\begin{axis}[ux axis,name=pa,width=\motW,height=52pt,
  xmin=0,xmax=70,ymin=0,ymax=52,
  xtick={0,10,...,70},ytick=\empty,
  y axis line style={draw=none}]
  \mlab{14.3}{45.8}{other}{ao}
  \mlab[text=motwarntext]{48.6}{45.8}{frozen path}{af}
  \mlab[anchor=base west,xshift=1.5pt]{0}{21.65}{small ops}{as}
  \mlab{17.7}{21.65}{LoRA}{al}
  \mlab{11.6}{3.5}{cast/copy}{ac}
  \mlab{22.6}{3.5}{GDN}{ag}
  \node[chart val] (av1) at (axis cs:68.6,35.3) {68.6};
  \node[chart val] (av2) at (axis cs:26.7,14.65) {26.7};
  \node[ux/label,inner xsep=0pt,inner ysep=1.5pt,anchor=base west] (aw) at ([xshift=2.5pt]av2.base east) {without frozen path};
  \mg{0}{0}{35.3}
  \fgrid{30}{{axis cs:0,0}}{av2.south}\fgrid{30}{av2.north}{{axis cs:0,35.3}}
  \fgrid{40}{{axis cs:0,0}}{aw.south}\fgrid{40}{aw.north}{{axis cs:0,35.3}}
  \fgrid{50}{{axis cs:0,0}}{aw.south}\fgrid{50}{aw.north}{{axis cs:0,35.3}}
  \fgrid{60}{{axis cs:0,0}}{aw.south}\fgrid{60}{aw.north}{{axis cs:0,35.3}}
  \fgrid{70}{{axis cs:0,0}}{av1.south}
  \uxhbar{35.3}{0}{28.6}{chart base}
  \uxhbar{35.3}{28.6}{68.6}{chart warn}
  \draw[ux/brace,draw=papergray] ([yshift=5pt]axis cs:0.3,35.3) -- ([yshift=5pt]axis cs:28.3,35.3);
  \draw[ux/brace,draw=motwarn] ([yshift=5pt]axis cs:28.9,35.3) -- ([yshift=5pt]axis cs:68.3,35.3);
  \uxhbar{14.65}{0}{9.1}{chart base}\uxhbar{14.65}{9.1}{15.3}{chart base}
  \uxhbar{14.65}{15.3}{20.1}{chart base}\uxhbar{14.65}{20.1}{24.4}{chart base}
  \uxhbar{14.65}{24.4}{26.7}{chart base}
\end{axis}}
\newcommand{\motname}[4][]{\node[font=\uxfont,inner xsep=0pt,inner ysep=1.5pt,
  anchor=base west,#1] (#4) at ([xshift=2.5pt]#2.base east) {#3};}
\newcommand{\motPanelB}[1][]{%
\begin{axis}[ux axis,name=pb,#1,width=\motW,height=46.15pt,
  xmin=0,xmax=30,ymin=0,ymax=46.15,
  xtick={0,5,...,30},ytick=\empty,
  y axis line style={draw=none}]
  \node[chart val] (bv1) at (axis cs:9.3,41.7) {9.3};
  \node[chart val] (bv2) at (axis cs:18.8,32.8) {18.8};
  \node[chart val] (bv3) at (axis cs:20.2,23.9) {20.2};
  \node[chart val ours] (bv4) at (axis cs:9.6,15.0) {9.6};
  \node[chart val] (bv5) at (axis cs:27,6.1) {98.0};
  \motname{bv1}{FP16}{bn1}\motname{bv2}{Q8 (inference)}{bn2}\motname{bv3}{+ gates FP16}{bn3}
  \motname[text=paperblue,font=\uxfont\bfseries]{bv4}{+ GDN FP16}{bn4}\motname{bv5}{Q4}{bn5}
  \mg{0}{0}{46.15}\mg{5}{0}{46.15}
  \fgrid{10}{{axis cs:0,0}}{bv4.south}\fgrid{10}{bv4.north}{bv1.south}\fgrid{10}{bv1.north}{{axis cs:0,46.15}}
  \fgrid{15}{{axis cs:0,0}}{bn4.south}\fgrid{15}{bn4.north}{{axis cs:0,46.15}}
  \fgrid{20}{{axis cs:0,0}}{bv2.south}\fgrid{20}{bv2.north}{{axis cs:0,46.15}}
  \fgrid{25}{{axis cs:0,0}}{bn3.south}\fgrid{25}{bn3.north}{bn2.south}\fgrid{25}{bn2.north}{{axis cs:0,46.15}}
  \fgrid{30}{bn5.north}{bn3.south}\fgrid{30}{bn2.north}{{axis cs:0,46.15}}
  \uxhbar{41.7}{0}{9.3}{chart base}
  \uxhbar{32.8}{0}{18.8}{chart warn}
  \uxhbar{23.9}{0}{20.2}{chart base}
  \uxhbar{15.0}{0}{9.6}{chart ours}
  \uxhbar{6.1}{0}{27}{chart base}
  \fill[white] ([yshift=-4.5pt]axis cs:22.6,6.1) -- ([yshift=-4.5pt]axis cs:23.2,6.1) -- ([yshift=4.5pt]axis cs:23.8,6.1)
    -- ([yshift=4.5pt]axis cs:23.2,6.1) -- cycle;
  \draw[papergray,line width=0.5pt] ([yshift=-4.5pt]axis cs:22.6,6.1) -- ([yshift=4.5pt]axis cs:23.2,6.1)
    ([yshift=-4.5pt]axis cs:23.2,6.1) -- ([yshift=4.5pt]axis cs:23.8,6.1);
  \draw[chart line ref] (axis cs:7.7,0) -- (axis cs:7.7,46.15);
\end{axis}}
\begin{tikzpicture}
\motPanelA
\ftitle{pa}{a}{Self CPU time (s)}
\motPanelB[at={(0,-26pt)},anchor=north west]
\ftitle{pb}{b}{Q3.5-4B gradient error (\%)}
\end{tikzpicture}
\vspace{-0.5em}
\caption{The frozen path dominates CPU training, and inference precision is not gradient precision. (a) Self CPU time of a training example on P2 with and without the frozen path (replaced by zeros), and the remaining time by operator; (b) LoRA-gradient error against FP32 by frozen-weight format, where + keeps the gate or the recurrent (GDN) projections of Q8 in FP16 (dark blue: our system's format; dashed: the error with BF16 weights).}
\label{fig:motivation}
\vspace{-2em}
\end{figure}
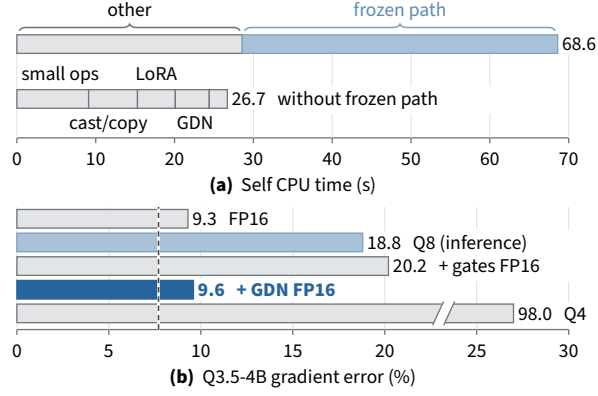

We profile a CPU LoRA trainer on an older phone, P1 (Qualcomm Snapdragon 8 Elite, Hexagon V79 NPU), and a newer one, P2 (Snapdragon 8 Elite Gen 5, Hexagon V81 NPU), both detailed in Section~\ref{sec:setup}. Even after basic CPU fixes, the frozen path takes most of the self CPU time of one training example on both phones (Figure~\ref{fig:motivation}a and Section~\ref{sec:speed}).

The NPU is built for this work: at the same forward shape, its matrix engine, Hexagon Matrix Extensions (HMX), far outpaces floating-point matrix multiplication on the CPU with Arm's Scalable Matrix Extension. Effective offload should also reduce the remaining CPU work: without the frozen path, small operators, casts and copies, adapter products, and GDN recurrence still take a large share of CPU time (Figure~\ref{fig:motivation}a).

\subsection{Numerical Precision in Backpropagation}\label{sec:precisionobs}

Inference weight formats are not automatically safe for backward. Figure~\ref{fig:motivation}b compares the relative $L_2$ error of concatenated LoRA gradients of Qwen3.5-4B (Q3.5-4B)~\cite{qwen35} under frozen-weight formats against a single-precision (FP32) reference in a harness that emulates the engine's arithmetic (Appendix~\ref{app:numerics}); its dashed line marks bfloat16 (BF16) weights. Q8, the engine's inference format, raises the error well above that of FP16 weights, and its 4-bit counterpart (Q4) destroys the gradient. The error concentrates in the recurrent projections: keeping only their gates in FP16 does not help, but restoring only the recurrent projection types to FP16 returns it to the FP16 level (Section~\ref{sec:fidelity}). We attribute this to the recurrent GDN state, which carries a weight perturbation into every later token.

\subsection{Limitations of the NPU Inference Interface}\label{sec:gap}

\looseness=-1 On both phones' NPUs, HMX works with Hexagon Vector Extensions (HVX), DMA, and an on-chip vector tightly coupled memory (VTCM). HMX accepts FP16 operands, accumulates in wider cells, and writes FP16 results~\cite{hmx}, so an FP32 CPU buffer cannot recover values lost in the engine's output conversion. The inference interface and its Q8 tiles serve forward products; backward traverses the other dimension, where a separately quantized transpose would add storage and change the effective matrix.

These findings require our system to (R1) reuse the forward weights and their dequantization arithmetic in backward, (R2) keep gradients usable within the engine's FP16 precision and range, and (R3) shrink and hide the CPU work and CPU--NPU boundary cost that offload exposes, for an hours-long job in the phone's memory and thermal budget.

\section{System Overview}\label{sec:overview}

To make first-order LoRA training practical on a phone, the system should accelerate the dominant frozen-weight computation through an inference-only NPU interface, preserve the gradients needed to train the adapters, and sustain execution within the phone's memory and thermal limits. We propose a CPU--NPU training framework that addresses these requirements by organizing execution around frozen projections. It combines a CPU trainer that owns the training graph and mutable state, a resident NPU service that supplies frozen-weight products in both directions, and a job controller that manages progress and pauses. This division lets forward, activation recomputation, and backpropagation share the NPU's resident weights throughout an update.

The framework delegates frozen-weight computation to the resident NPU service while the CPU trainer retains control of training. By extending the inference engine to support backpropagation with the same resident weights, the service enables NPU acceleration across the frozen path without an additional weight copy. Numerical safeguards preserve usable gradients, allowing the CPU trainer to learn the adapters from the service's results.
The CPU--NPU runtime organizes the trainer and service into a coordinated training process. It reduces the remaining CPU work and overlaps independent computation across the two processors, helping translate NPU acceleration into faster training steps. It also manages activation memory so training can fit within the phone's available memory.

The job controller extends this execution model across an hours-long update. At optimizer-step boundaries, when no NPU request is in flight, it saves the CPU-owned training state so execution can resume after a pause or service restart. Foreground use and thermal limits can therefore interrupt training without losing committed progress.

\section{Frozen-Path Backpropagation}\label{sec:backward}

\begin{figure*}[t]
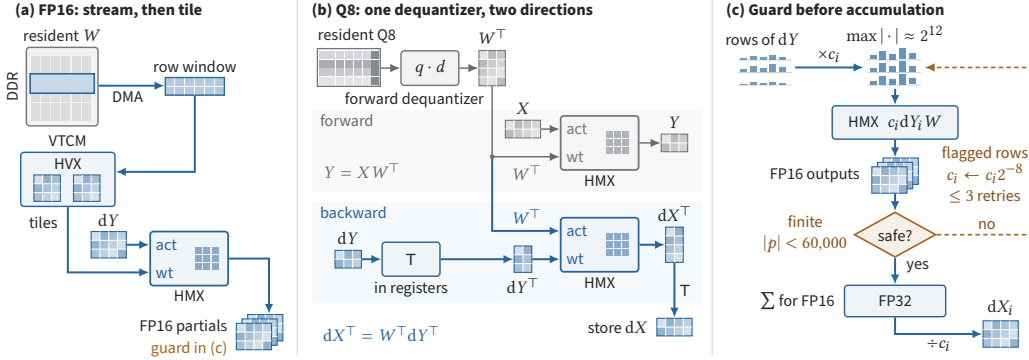

\centering
\begingroup
\tikzset{
  kr/text/.style={font=\uxfont,inner sep=1pt,align=center,text=black!85},
  kr/title/.style={kr/text,font=\uxfont\bfseries,anchor=west},
  kr/flow/.style={-{UxTip},draw=paperblue,line width=0.9pt},
  kr/fwd/.style={-{UxTip},draw=papergray!90!black,line width=0.65pt},
  kr/retry/.style={-{UxTip},draw=paperorange,line width=0.75pt,
    dash pattern=on 2pt off 1.5pt},
  kr/box/.style={draw=paperblue,fill=paperblue!7,rounded corners=1.5pt,
    line width=0.65pt},
  kr/neutral/.style={draw=papergray,fill=paperlight,rounded corners=1.5pt,
    line width=0.65pt},
  pics/kr/matrix/.style n args={3}{code={
    \pgfmathtruncatemacro\krmaxr{#2-1}
    \pgfmathtruncatemacro\krmaxc{#3-1}
    \foreach \r in {0,...,\krmaxr}{\foreach \c in {0,...,\krmaxc}{
      \pgfmathtruncatemacro\krshade{20+mod(\r*7+\c*11,4)*10}
      \fill[#1!\krshade] ({(\c-#3/2)*0.14},{(#2/2-\r-1)*0.14}) rectangle ++(0.14,0.14);}}
    \draw[#1,line width=0.5pt] ({-#3*0.07},{-#2*0.07}) rectangle ({#3*0.07},{#2*0.07});
    \foreach \r in {1,...,\krmaxr}{\draw[white,line width=0.35pt]
      ({-#3*0.07},{#2*0.07-\r*0.14}) -- ({#3*0.07},{#2*0.07-\r*0.14});}
    \foreach \c in {1,...,\krmaxc}{\draw[white,line width=0.35pt]
      ({-#3*0.07+\c*0.14},{-#2*0.07}) -- ({-#3*0.07+\c*0.14},{#2*0.07});}
  }},
  pics/kr/hmx/.style args={#1}{code={
    \draw[draw=#1,fill=#1!7,line width=0.8pt,rounded corners=1.5pt]
      (-0.62,-0.43) rectangle (0.62,0.43);
    \node[kr/text,anchor=west,text=#1] at (-0.57,0.22) {act};
    \node[kr/text,anchor=west,text=#1] at (-0.57,-0.22) {wt};
    \foreach \r in {0,...,2}{\foreach \c in {0,...,2}{
      \fill[#1!55] ({0.10+\c*0.13},{-0.18+\r*0.13}) rectangle ++(0.105,0.105);}}
    \node[kr/text,anchor=north] at (0,-0.47) {HMX};
  }},
}
\resizebox{\textwidth}{!}{%
\begin{tikzpicture}[x=1cm,y=1cm]
\path[use as bounding box] (0,0) rectangle (16.8,5.8);
\draw[papergray!30,line width=0.5pt] (4.65,0.08) -- (4.65,5.75)
  (11.25,0.08) -- (11.25,5.75);
\input{figures/redraw/kernels-f16}
\input{figures/redraw/kernels-q8}
\input{figures/redraw/kernels-range}
\end{tikzpicture}}%
\endgroup

\vspace{-0.5em}
\caption{Input gradients reuse resident weights, FP16 in sensitive ones such as recurrent projections and Q8 in the rest. (a) FP16 row windows stream into VTCM, and HVX converts them into HMX tiles; (b) the Q8 kernel reuses the forward dequantizer, feeds $W^\top$ to the activation port, and transposes gradients in registers and on store; (c) the client scales rows, kernels flag unsafe FP16 outputs before the FP16 kernel sums FP16 partials in FP32, only flagged rows retry with a smaller scale, and the client unscales successful results.}
\label{fig:kernels}

\end{figure*}

On the NPU, there are two tasks in the frozen-path backpropagation: compute $\dy W$ from the resident weight tiles and keep it usable under quantized weights and FP16 outputs.

\subsection{A projection-level interface}

The NPU's inference API runs forward graphs, but a training graph is dynamic, may contain recurrent layers, and owns an optimizer; porting it to the NPU's compiler would mean rewriting all three. The service instead exposes one operation per frozen projection with a direction flag: forward ($XW^\top$) or input gradient ($\dy W$, Equation~\eqref{eq:dx}), so the graph and its optimizer never leave the CPU. The client runs the device's kernel planner, so it knows at startup which projection shapes each direction supports on the NPU. Shapes outside that set, such as gate projections with only a few outputs, run instead in an explicit floating-point CPU implementation that returns complete outputs and gradients.\label{sec:interface}

\subsection{Input gradients from resident FP16 weights}\label{sec:f16}

Backward traverses the other dimension of $W$. A transposed FP16 copy for it would double resident storage, so our FP16 backward kernel reads $W$ in place and transposes it on chip, one window at a time (Figure~\ref{fig:kernels}a). DMA streams row windows of the resident $W$ into VTCM, HVX converts each window into the tiles that the transposed product needs, and the output gradient takes the engine's activation conversion path. The kernel planner chooses reduction segments, window width, and token tiles within the VTCM budget.

Segmenting also decides where the result is rounded. Large projections split the contraction dimension into segments, and each segment yields an FP16 partial result that HVX expands and accumulates in FP32. Because each partial result is rounded before the FP32 sum, later cancellation cannot repair an overflowed partial result, so the range guard detects it before accumulation and resends the row (Section~\ref{sec:range}).

\subsection{Operand exchange for Q8 tiles}\label{sec:q8}

For Q8, backward uses forward's effective matrix by reusing both the quantized values and the forward dequantizer's arithmetic. A separately quantized transposed copy would change the values, and even a separate conversion of the same integers changes the effective matrix, because the vector dequantizer can round a quantized value $q$ with scale $d$ differently from a conventional FP16 rounding of $qd$. We therefore exchange the two HMX operand roles (Figure~\ref{fig:kernels}b): the dequantized weight tile becomes the activation operand and the output-gradient tile the weight operand, which computes
\begin{equation}\label{eq:exchange}
(\dx_{\rm base})^\top=W^\top\dy^\top.
\end{equation}
The forward dequantizer already produces the FP16 layout the $W^\top$ activation tile needs, so only the gradients are transposed: the kernel transposes $\dy$ in registers into the weight layout and each output tile as it stores $\dx$. The logical transpose of Equation~\eqref{eq:exchange} thus happens at tile boundaries, without a full FP16 expansion or a second quantized model, so backward reads forward's effective weights (Section~\ref{sec:fidelity}).

\begin{figure*}[t]
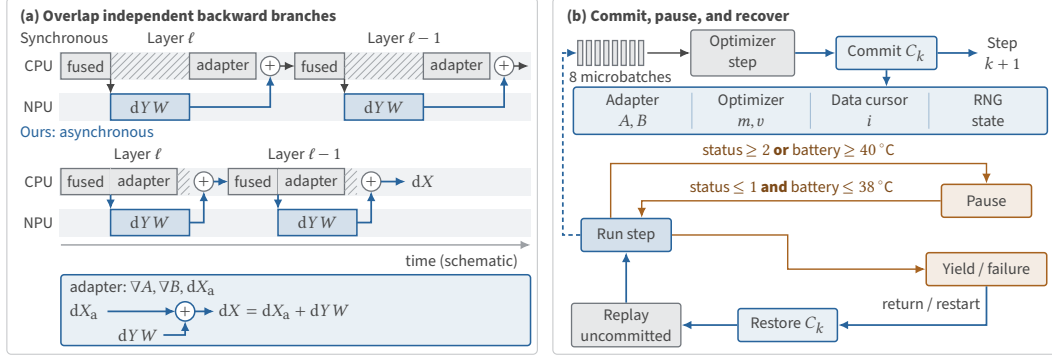

\centering
\resizebox{\textwidth}{!}{%
\begin{tikzpicture}[x=1cm,y=1cm,
  r4/text/.style={font=\uxfont,inner sep=1pt,align=center,text=uxink},
  r4/title/.style={r4/text,font=\uxfont\bfseries,anchor=west},
  r4/cpu/.style={draw=papergray,fill=papergray!17,line width=.6pt},
  r4/npu/.style={draw=paperblue,fill=paperblue!20,line width=.7pt},
  r4/commit/.style={draw=paperblue,fill=paperblue!10,line width=.7pt},
  r4/guard/.style={draw=paperorange,fill=paperorange!12,line width=.7pt},
  r4/flow/.style={-{UxTip},draw=uxink,line width=.65pt},
  r4/newflow/.style={r4/flow,draw=paperblue,line width=.8pt},
  r4/guardflow/.style={r4/flow,draw=paperorange},
  r4/box/.style={rounded corners=1.3pt,minimum height=.53cm,inner sep=3pt},
  r4/panel/.style={draw=papergray!40,fill=white,line width=.5pt,rounded corners=2pt}]
  \path[use as bounding box] (0,0) rectangle (17,5.8);
  \draw[r4/panel] (.03,.03) rectangle (8.57,5.77);
  \draw[r4/panel] (8.82,.03) rectangle (16.97,5.77);
  \input{figures/redraw/runtime-schedule}
  \begin{scope}[shift={(9.0,0)}]
    \node[r4/title] at (0,5.47) {\textbf{(b)} Commit, pause, and recover};
    \input{figures/redraw/runtime-commit}
    \input{figures/redraw/runtime-lifecycle}
  \end{scope}
\end{tikzpicture}}%
\vspace{-1em}
\caption{Our CPU--NPU training runtime. (a) Schematic two-layer schedules: after fused CPU work, asynchronous requests overlap CPU adapter gradients with NPU $\dy W$, both branches join before the next layer, and hatching marks exposed CPU wait; (b) the dashed arrow expands one optimizer step of eight accumulated microbatches, whose commit saves the complete restart state; thermal pause and resume use hysteresis, and a yield or failure restores the last commit and replays only the uncommitted training work.}
\label{fig:runtime}
\end{figure*}

\subsection{A gradient-sensitive precision policy}\label{sec:policy}

Reusing the forward dequantizer makes forward and backward use the same effective weights, but it does not fix Q8's gradient error. Section~\ref{sec:precisionobs} traces that error to projections whose quantization error a recurrent state carries forward, not to gates alone. Keeping every projection in FP16 would increase resident weight storage, so our policy sets precision per projection type from measured gradient sensitivity: FP16 for the sensitive types, such as all recurrent projection types where a model has them, and Q8 for the rest, such as attention and feed-forward (MLP) projections; attention-only models use Q8 throughout. Activations stay in floating point when weights are quantized, separating storage and computation precision as QLoRA does~\cite{qlora}. Because a CPU Q8 path can quantize activations, startup validation keeps every shape off it; unsupported shapes take the floating-point CPU implementation of Section~\ref{sec:interface}.

\subsection{Range guard at the request boundary}\label{sec:range}

The engine's FP16 outputs limit range at both ends: small gradients underflow, and large partial results overflow. A global loss scale~\cite{mixedprecision} cannot serve rows whose magnitudes differ widely, so the client scales each row by a power of two, which adds no rounding (Figure~\ref{fig:kernels}c). For each finite, nonzero row $\dy_i$, the client chooses a factor $c_i$ that places $\max_j|c_i\dy_{ij}|$ near $2^{12}$; the service computes the scaled product, and the client divides its FP32 result by $c_i$. Since $c_i$ cannot bound every FP16 partial result, the kernel flags rows whose partial result is nonfinite or reaches 60,000, below FP16's maximum of 65,504, and the client resends only those rows, scaling $c_i$ by $2^{-8}$ per retry, up to three times.

\section{A CPU--NPU Training Runtime}\label{sec:runtime}

The NPU service of Section~\ref{sec:backward} supplies frozen-path outputs and input gradients; the CPU retains the surrounding computation and training state. The runtime organizes this division into complete training steps by reducing CPU work and scheduling projection requests and activation recomputation within a memory budget. Completed optimizer steps provide boundaries for saving state, allowing training to pause and resume under changing phone conditions.

\subsection{Reducing CPU work between NPU requests}\label{sec:fusion}

GDN recurrence and LoRA branches contain many small operations, and the vocabulary head also remains on the CPU after offload (Figure~\ref{fig:motivation}a). We fuse each into a separate autograd node, keeping frozen projections outside for NPU execution. The GDN node combines causal convolution, SiLU, normalization, gates, delta update, and gated RMS normalization between its projections. It keeps recurrent state in FP32 while preserving BF16 rounding boundaries. Internal state checkpoints every 16 tokens let backward replay each segment forward and differentiate it in reverse. Partitioning by value head and 64-column slices exposes CPU parallelism without retaining per-token intermediates.

Each LoRA node combines rank-16 products, dropout, scaling, and addition. Its separation from the frozen projection allows overlap with NPU execution. With answer-only supervision, the vocabulary head computes logits only at supervised answer positions; we fuse its BF16 matrix path for forward and backward and keep logits in FP32.

\subsection{Scheduling training with memory budget}\label{sec:overlap}\label{sec:memory}

Resident weights, training state, and foreground apps share the phone's memory, so retaining every layer's activations can leave insufficient space for backward temporaries. The trainer therefore bounds activation storage by the smaller of a configured cap and currently available memory after reserving space for those temporaries. Layers whose activations cannot be retained within this budget are recomputed through the same NPU projection interface. The resulting schedule follows the training graph's dependencies and includes the forward requests needed for recomputation.

Within this schedule, the LoRA and frozen branches can execute independently until their results are combined. In backward, the client submits $\dy W$, computes adapter gradients, and waits only to add the branches' input gradients (Figure~\ref{fig:runtime}a). In forward, it submits projections sharing an input ahead of use. The fused LoRA node thus supplies CPU work to overlap with NPU requests.

One request is in flight at a time, and every result, including recomputation results, is consumed exactly once at its intended graph position. After the accumulated microbatches, the CPU updates the adapters with no request outstanding, establishing the boundary for progress saving.

\subsection{Pausing and resuming across training steps}\label{sec:job}

To support pauses and restarts, the job saves CPU-owned training state at optimizer-step boundaries, when no NPU request is outstanding. This includes the adapter parameters, optimizer state, etc. On resumption, the trainer restores the last commit and repeats only uncommitted work; a supervisor restarts the NPU service if it has failed (Figure~\ref{fig:runtime}b).

To manage heat and limit interference with foreground use, the trainer can run at lower priority with threads pinned to a subset of cores. A thermal controller monitors Android's thermal status and battery temperature. Training pauses when either reaches its pause threshold and resumes only when both fall to or below their lower resume thresholds.

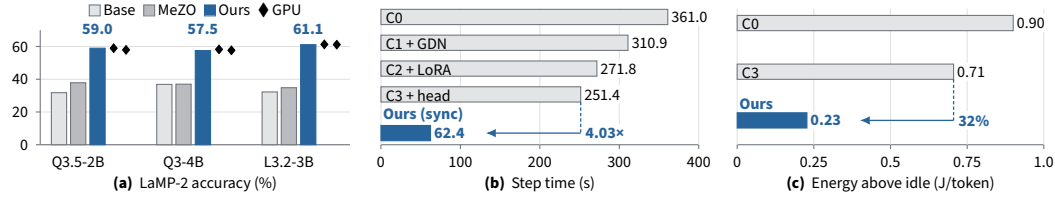
\begin{figure*}[t]
\centering
\tikzset{
  chart base/.style={fill=papergray!20,draw=papergray,line width=0.5pt},
  chart base2/.style={fill=papergray!50,draw=papergray,line width=0.5pt},
  chart ours/.style={fill=paperblue,draw=paperblue,line width=0.5pt},
  chart warn/.style={fill=paperorange!50,draw=paperorange,line width=0.5pt},
  chart band/.style={fill=paperorange!15,draw=none},
  chart line ours/.style={draw=paperblue,line width=1pt},
  chart line base/.style={draw=papergray,line width=1pt,dash pattern=on 2.5pt off 1.5pt},
  chart line ref/.style={draw=uxink,line width=0.6pt,dash pattern=on 2pt off 1.5pt,
    preaction={draw=white,line width=1pt,solid}},
  chart val/.style={ux/value,anchor=west},
  chart val ours/.style={ux/value,anchor=west,text=paperblue,font=\uxfont\bfseries},
  chart title/.style={font=\uxfont,inner sep=0pt,anchor=base},
}
\newcommand{\ftitledrop}{20pt}
\newcommand{\ftitle}[3]{\node[chart title] at ([yshift=-\ftitledrop]#1.south)
  {\textbf{(#2)}\hspace{3pt}#3};}
\newcommand{\chartkey}[2]{{\setlength{\fboxsep}{0pt}\setlength{\fboxrule}{0.5pt}%
  \fcolorbox{#2}{#1}{\rule{0pt}{4.5pt}\rule{4.5pt}{0pt}}}}
\newcommand{\fgrid}[3]{\draw[/pgfplots/every axis grid]
  ({axis cs:#1,0} |- #2) -- ({axis cs:#1,0} |- #3);}
\newcommand{\hlH}{1.85cm}
\newcommand{\hlHr}{2.2cm}
\newcommand{\hlPanelA}[1][]{%
\begin{axis}[ux axis,name=ha,#1,width=5.0cm,height=\hlH,
  xmin=0.64,xmax=3.58,ymin=0,ymax=70,
  xtick={1,2,3},xticklabels={Q3.5-2B,Q3-4B,L3.2-3B},xtick style={draw=none},
  x axis line style={draw=papergray},
  ytick={0,20,40,60},ymajorgrids=true]
  \uxvbar{1}{-8.8pt}{31.79}{chart base}
  \uxvbar{1}{0pt}{37.86}{chart base2}
  \uxvbar{1}{8.9pt}{58.96}{chart ours}
  \uxvbar{2}{-8.8pt}{36.85}{chart base}
  \uxvbar{2}{0pt}{36.99}{chart base2}
  \uxvbar{2}{8.9pt}{57.51}{chart ours}
  \uxvbar{3}{-8.8pt}{32.23}{chart base}
  \uxvbar{3}{0pt}{34.83}{chart base2}
  \uxvbar{3}{8.9pt}{61.13}{chart ours}
  \node[ux/refmark] at ([xshift=16.3pt]axis cs:1,58.96) {};
  \node[ux/refmark] at ([xshift=22pt]axis cs:1,57.95) {};
  \node[ux/refmark] at ([xshift=16.3pt]axis cs:2,58.24) {};
  \node[ux/refmark] at ([xshift=22pt]axis cs:2,57.66) {};
  \node[ux/refmark] at ([xshift=16.3pt]axis cs:3,61.27) {};
  \node[ux/refmark] at ([xshift=22pt]axis cs:3,61.13) {};
  \node[chart val ours,anchor=south] at ([xshift=8.9pt]axis cs:1,65.0) {59.0};
  \node[chart val ours,anchor=south] at ([xshift=8.9pt]axis cs:2,65.0) {57.5};
  \node[chart val ours,anchor=south] at ([xshift=8.9pt]axis cs:3,65.0) {61.1};
\end{axis}}
\newcommand{\hlin}[2]{\node[ux/label,anchor=west,inner xsep=2pt,inner ysep=0pt,
  text height=5.6pt,text depth=0pt] at (axis cs:0,#1) {#2};}
\newcommand{\hlours}[3]{\node[ux/label,anchor=south west,inner xsep=0.8pt,
  inner ysep=1.5pt,font=\uxfont\bfseries,text=paperblue] (#3) at ([yshift=\uxbarw]axis cs:0,#1) {#2};}
\newcommand{\hlv}[4]{\node[chart val] (#4) at (axis cs:#1,#2) {#3};}
\newcommand{\hlbot}{{rel axis cs:0,0}}
\newcommand{\hltop}{{rel axis cs:0,1}}
\newcommand{\hlPanelB}[1][]{%
\begin{axis}[ux axis,name=hb,#1,width=5.13cm,height=\hlHr,
  xmin=0,xmax=400,ymin=-0.95,ymax=4.33,
  xtick={0,100,200,300,400},ytick=\empty,
  y axis line style={draw=none}]
  \hlv{361.0}{4}{361.0}{hb4}\hlv{310.9}{3}{310.9}{hb3}
  \hlv{271.8}{2}{271.8}{hb2}\hlv{251.4}{1}{251.4}{hb1}
  \hlours{-0.5}{Ours (sync)}{hbn}
  \node[chart val ours] (hbv) at (axis cs:62.4,-0.5) {62.4};
  \node[ux/label,anchor=west,xshift=1pt,inner ysep=1.5pt,text=paperblue,
    font=\uxfont\bfseries] (hbr) at (axis cs:251.4,-0.5) {4.03\texttimes};
  \fgrid{0}{\hlbot}{\hltop}\fgrid{200}{\hlbot}{\hltop}
  \fgrid{100}{hbn.north}{\hltop}
  \fgrid{300}{hbr.north}{hb1.south}\fgrid{300}{hb1.north}{hb2.south}
  \fgrid{300}{hb2.north}{\hltop}
  \fgrid{400}{\hlbot}{hb4.south}
  \uxhbar{4}{0}{361.0}{chart base}
  \uxhbar{3}{0}{310.9}{chart base}
  \uxhbar{2}{0}{271.8}{chart base}
  \uxhbar{1}{0}{251.4}{chart base}
  \uxhbar{-0.5}{0}{62.4}{chart ours}
  \hlin{4}{C0}\hlin{3}{C1 + GDN}\hlin{2}{C2 + LoRA}\hlin{1}{C3 + head}
  \draw[paperblue,line width=0.6pt,dash pattern=on 1.5pt off 1pt]
    ([yshift=-3.5pt]axis cs:251.4,1) -- (axis cs:251.4,-0.5);
  \draw[-{UxTip},paperblue,line width=0.6pt] (axis cs:251.4,-0.5) -- (axis cs:132,-0.5);
\end{axis}}
\newcommand{\hlPanelC}[1][]{%
\begin{axis}[ux axis,name=hc,#1,width=4.95cm,height=\hlHr,
  xmin=0,xmax=1,ymin=-0.5,ymax=2.3,
  xtick={0,0.25,0.5,0.75,1},xticklabels={0,0.25,0.5,0.75,1.0},ytick=\empty,
  y axis line style={draw=none}]
  \hlv{0.9002}{2}{0.90}{hc2}\hlv{0.7057}{1}{0.71}{hc1}
  \hlours{0}{Ours}{hcn}
  \node[chart val ours] (hcv) at (axis cs:0.2275,0) {0.23};
  \node[ux/label,anchor=west,xshift=1pt,inner ysep=1.5pt,text=paperblue,
    font=\uxfont\bfseries] (hcr) at (axis cs:0.7057,0) {32\%};
  \fgrid{0}{\hlbot}{\hltop}\fgrid{0.5}{\hlbot}{\hltop}
  \fgrid{0.25}{\hlbot}{hcv.south}\fgrid{0.25}{hcv.north}{\hltop}
  \fgrid{0.75}{\hlbot}{hcr.south}\fgrid{0.75}{hcr.north}{hc1.south}
  \fgrid{0.75}{hc1.north}{\hltop}
  \fgrid{1}{\hlbot}{hc2.south}\fgrid{1}{hc2.north}{\hltop}
  \uxhbar{2}{0}{0.9002}{chart base}
  \uxhbar{1}{0}{0.7057}{chart base}
  \uxhbar{0}{0}{0.2275}{chart ours}
  \hlin{2}{C0}\hlin{1}{C3}
  \draw[paperblue,line width=0.6pt,dash pattern=on 1.5pt off 1pt]
    ([yshift=-3.5pt]axis cs:0.7057,1) -- (axis cs:0.7057,0);
  \draw[-{UxTip},paperblue,line width=0.6pt] (axis cs:0.7057,0) -- (axis cs:0.4,0);
\end{axis}}
\resizebox{\textwidth}{!}{%
\begin{tikzpicture}
\hlPanelA[at={(0,0)},anchor=south west]
\hlPanelB[at={(5.5cm,0)},anchor=south west]
\hlPanelC[at={(11.25cm,0)},anchor=south west]
\node[ux/ptitle,anchor=base] at ([yshift=6.8pt]ha.north) {%
  \chartkey{papergray!20}{papergray}\,Base\hspace{3.5pt}%
  \chartkey{papergray!50}{papergray}\,MeZO\hspace{3.5pt}%
  \chartkey{paperblue}{paperblue}\,Ours\hspace{3.5pt}%
  \uxdiamond\,GPU};
\ftitle{ha}{a}{LaMP-2 accuracy (\%)}
\ftitle{hb}{b}{Step time (s)}
\ftitle{hc}{c}{Energy above idle (J/token)}
\end{tikzpicture}}%
\caption{Our phone-trained adapters approach GPU-trained ones, and our steps are faster and use less energy than the optimized CPU trainer's. (a) LaMP-2 accuracy (diamonds: two GPU-trained seeds); (b) step time on P1 (screen off) and (c) device energy above idle per token on P2 (battery, screen on), both cooled and on one 3,431-token batch; arrows give our speedup over C3 and our share of C3's energy.}
\label{fig:headline}
\vspace{-1em}
\end{figure*}

\section{Implementation}\label{sec:impl}

\noindent\textbf{Software stack.}
The CPU trainer uses PyTorch 2.12.1~\cite{pytorch} and CPython 3.12 in user-space Linux. Arm C++ extensions implement fused operators; a custom \texttt{autograd.Function} connects frozen projections to the NPU service. The service extends a modified llama.cpp ggml Hexagon backend~\cite{ggmlhexagon}; its Q8\_0 and Q4\_0 formats correspond to Q8 and Q4. FP16 backward builds on ggml's \texttt{OUT\_PROD}, while added Q8 kernels implement operand exchange.

\noindent\textbf{Startup validation.}
Client and device share one kernel planner. Startup checks validate supported shapes, token buckets, and directions against a reference, probe determinism and overflow, and exclude ggml's CPU Q8 path, which quantizes activations. Graphs and tensor metadata are then reused throughout training.

\noindent\textbf{Model integration.}
Supported models are Q3.5-4B (248 projections), Qwen3.5-2B (Q3.5-2B), Qwen3.5-0.8B (\mbox{Q3.5-0.8B}), Qwen3-4B~\cite{qwen3} (Q3-4B; 252), and \mbox{Llama-3.2-3B}~\cite{llama32} (L3.2-3B; 196). Smaller Qwen3.5 models run 16-output gates on the CPU with complete forward and backward. We integrate a model by supplying its projection list, checkpoint names, CPU operators for non-projection layers, and precision configuration.

\noindent\textbf{Runtime configuration.}
Nightly updates use a 3000\,MB activation cap and 4000\,MB backward reserve. Thermal pacing pauses at Android thermal status 2 (moderate) or battery temperature 40\,$^\circ$C, resuming only at status 1 or below and battery temperature at most 38\,$^\circ$C.

\section{Evaluation}\label{sec:eval}

We evaluate the system through six research questions:
\begin{itemize}[leftmargin=*,itemsep=1pt,topsep=2pt,parsep=0pt]
\item \textbf{RQ1:} Can the system complete on-device fine-tuning with task quality comparable to GPU-trained references?
\item \textbf{RQ2:} How do NPU offload, CPU fusion, and asynchronous execution contribute to training performance?
\item \textbf{RQ3:} How do inference weight formats and FP16 arithmetic affect numerical fidelity?
\item \textbf{RQ4:} What energy and memory costs does training incur?
\item \textbf{RQ5:} Can training sustain progress under thermal limits and resume correctly after pauses or service failures?
\item \textbf{RQ6:} How does the system compare with phone-GPU training and forward-only training methods?
\end{itemize}
We first describe the experimental setup and then address each question in turn.

\subsection{Setup}\label{sec:setup}

\noindent\textbf{Platforms.}
P1 is a 16\,GB OnePlus 13T and P2 a 16\,GB OnePlus 15, with Adreno 830 and 840 GPUs. GPU references use an NVIDIA RTX 5090. Table~\ref{tab:setup} lists each experiment's conditions; the screen wakes during parts of both P1 overlap runs of the nightly update (Appendix~\ref{app:energy}).

\noindent\textbf{Models and workloads.}

We evaluate hybrid Q3.5-4B and Q3.5-2B and attention-only Q3-4B and L3.2-3B, with Q3.5-0.8B for scaling tests. The workloads cover a nightly personalization update using on-phone labels and the public LaMP-2 movie-tagging benchmark~\cite{lamp}. Both use first-order LoRA with answer-only supervision; the nightly update continues an existing adapter, while LaMP-2 trains a fresh one. Appendix~\ref{app:setup} details the models, datasets, and training settings.

\begin{table}[t]
\caption{Run conditions of each experiment on P1 and P2.}
\label{tab:setup}
\centering\tabstyle
\begin{tabular}{@{}lllllc@{}}
\toprule
Experiment & Phone & Model & Cooler & Screen & Threads\\
\midrule
Nightly update & P1 & Q3.5-4B & \icCool{} yes & \icScreenOff{} off & 8\\
Nightly update & P2 & Q3.5-4B & \icCool{} yes & \icScreenOff{} off & 6, 8\\
LaMP-2 epoch & P1 & Q3.5-2B & \icCool{} yes & \icScreenOff{} off & 8\\
LaMP-2 epoch & P2 & Q3-4B & \icCool{} yes & \icScreenOn{} on & 8\\
LaMP-2 epoch & P1 & L3.2-3B & \icCool{} yes & \icScreenOff{} off & 8\\
Controlled step & P1 & Q3.5-4B & \icCool{} yes & \icScreenOff{} off & 8\\
Energy & P2 & Q3.5-4B & \icCool{} yes & \icScreenOn{} on & 8\\
Thermal pacing & P2 & Q3.5-4B & \icNoCool{} no & \icScreenOff{} off & 6\\
Phone GPU & P1 & L3.2-3B & \icCool{} yes & \icScreenOff{} off & 8\\
Phone GPU & P2 & L3.2-3B & \icCool{} yes & \icScreenOn{} on & 8\\
\bottomrule
\end{tabular}
\vspace{-2em}
\end{table}

\begin{table}[t]
\caption{Complete on-device training runs on P1 and P2.}
\label{tab:runs}
\centering\tabstyle
\begin{tabular}{@{}cllrccrr@{}}
\toprule
 & Phone & Model & Tok. & Over- & Thr. & Time & Rate\\
 & & & (M) & lap & & (h) & (tok/s)\\
\midrule
\multirow{6}{*}{\rotatebox[origin=c]{90}{Nightly}}
 & P1 & Q3.5-4B & 1.48 & $\times$ & 8 & 8.51 & 48.4\\
 & P1 & Q3.5-4B & 1.48 & $\surd$ & 8 & 7.82 & 52.7\\
 & P1 & Q3.5-4B & 1.48 & $\surd$ & 8 & 7.87 & 52.3\\
 & P2 & Q3.5-4B & 1.48 & $\times$ & 8 & 9.01 & 45.7\\
 & P2 & Q3.5-4B & 1.48 & $\surd$ & 8 & 8.57 & 48.1\\
 & P2 & Q3.5-4B & 1.48 & $\surd$ & 6 & 8.09 & 51.0\\
\midrule  
\multirow{3}{*}{\rotatebox[origin=c]{90}{LaMP-2}}
 & P1 & Q3.5-2B & 1.23 & $\times$ & 8 & 3.09 & 110.5\\
 & P2 & Q3-4B & 1.23 & $\times$ & 8 & 9.06 & 37.6\\
 & P1 & L3.2-3B & 1.31 & $\times$ & 8 & 4.99 & 72.9\\
\bottomrule
\end{tabular}
\vspace{-2em}
\end{table}

\noindent\textbf{Baselines.}
To attribute performance gains, we compare CPU configurations C0--C3. C0 applies four basic optimizations: logits only at supervised positions, parallel dropout, eager attention, and faster convolution. C1--C3 successively add GDN, LoRA, and vocabulary-head fusion. C3 serves as the optimized CPU reference for evaluating NPU offload.

To compare alternative training approaches, we use QVAC Fabric~\cite{qvac} (Fabric; v10549.5.1, Vulkan) for phone-GPU LoRA training, and MeZO~\cite{mezo} for forward-only updates. MeZO's task-quality comparison runs on the RTX 5090 with the same LoRA targets and data, dropout 0, and a learning-rate sweep.

For task-quality references, we train GPU adapters with the same recipe across four seeds for the nightly update and two for LaMP-2. We compare against this spread because the fused LoRA operator draws its own dropout masks.

\noindent\textbf{Metrics.}
Training time counts only active optimizer steps, not loading, validation, or pauses. Throughput counts training tokens, including prompts. Energy is whole-device energy above idle on battery power. Kernel error is normalized mean squared error (NMSE). All LaMP-2 adapters are evaluated on the RTX 5090 with the BF16 base model, using greedy tags, accuracy, macro-F1, and exact paired McNemar tests. The nightly update is evaluated on a 175-item next-day set and a 510-item held-out set (Appendix~\ref{app:quality}).

\begin{table}[t]
\caption{Numerical fidelity from kernels to task quality.}
\label{tab:fidelity}
\centering\tabstyle
\begin{tabular}{@{}lrr@{}}
\toprule
Check & Ours & Reference\\
\midrule
Kernel NMSE, FP16 & $7.7{\times}10^{-8}$ & $<10^{-6}$\\
Kernel NMSE, Q8 & $1.2{\times}10^{-7}$ & $<10^{-6}$\\
One-hot effective weights & bit-identical & forward\\
Forward \texttt{MUL\_MAT} cases & 759/759 & base build\\
\midrule
Gradient error (\%), BF16\,/\,CPU Q8 & 10.15 & 10.9\,/\,21.1\\
Fused-GDN gradient error (\%) & 8.8 & 10.9\\
\midrule
Mean step-loss gap to GPU & 0.0050 & 0.0058\\
Steps in four-seed range (\%) & 65.8 & 60.0\\
\midrule
Next-day accuracy (\%) & 58.9 & 61.7\\
Held-out correct (of 510) & 241 & 243\\
\bottomrule
\end{tabular}
\vspace{-1.5em}
\end{table}

\begin{table}[t]
\caption{LaMP-2 development accuracy (\%) on 692 records.}
\label{tab:lamp_full}
\centering\tabstyle
\begin{tabular}{@{}lrrrrr@{}}
\toprule
Model & Base & MeZO & Ours & GPU 0 & GPU 1\\
\midrule
Q3.5-2B & 31.79 & 37.86 & 58.96 & 58.96 & 57.95\\
Q3-4B & 36.85 & 36.99 & 57.51 & 58.24 & 57.66\\
L3.2-3B & 32.23 & 34.83 & 61.13 & 61.27 & 61.13\\
\bottomrule
\end{tabular}
\vspace{-1.5em}
\end{table}

\subsection{End-to-end on-device fine-tuning (RQ1)}\label{sec:e2e}

RQ1 asks whether a complete update finishes on the phone and yields an adapter as useful as a GPU-trained one.

\noindent\textbf{Completion.}
Both phones complete all 222 optimizer steps of the nightly update (Table~\ref{tab:runs}). With overlap, the update takes 7.82 and 7.87 active hours in two P1 runs and 8.09 on P2 with six pinned threads. The system also trains a fresh adapter for a full LaMP-2 epoch on all three models, taking 3.09 to 9.06 active hours depending on model size.

\noindent\textbf{Reproducibility.}
All six nightly runs, across both phones and three runtime configurations, produce bit-identical losses at every step and the same final adapter, with no service restarts, row retries, or fused-operator fallbacks. Switching phones, thread counts, or overlap leaves each step's computation unchanged; overlap changes only when each product runs, not the result it returns (Section~\ref{sec:overlap}).

\noindent\textbf{Adapter quality.}
The phone-trained adapters are close to their GPU-trained references and far better than the base model. On LaMP-2 (Table~\ref{tab:lamp_full}), they land within 1.01 percentage points of GPU-trained adapters on all three models and gain 21 to 29 points over the base model (Figure~\ref{fig:headline}a). They also match the GPU seeds in macro-F1 (Table~\ref{tab:lamp_f1}), exceeding both seeds on Q3.5-2B and L3.2-3B and staying close on Q3-4B. Exact paired McNemar tests of our adapters against the two GPU seeds give $p=1.00$ and $0.23$ on \mbox{Q3.5-2B}, $0.38$ and $1.00$ on Q3-4B, and $1.00$ for both on L3.2-3B, so no model's adapter differs significantly from either seed. On the 175-item next-day set, the phone-trained nightly adapter answers 103 items correctly (58.9\%), against 108 (61.7\%) for a same-recipe GPU reference ($p=0.23$; Table~\ref{tab:fidelity}).

\subsection{Where the speed comes from (RQ2)}\label{sec:speed}

\noindent\textbf{Offload after CPU optimization.}
We first ask whether the NPU still helps after CPU optimization. We run every configuration on one 3,431-token batch on P1 and add one mechanism at a time (Figure~\ref{fig:headline}b). Fusing GDN, LoRA, and the head shortens the CPU step from 361.0 to 251.4\,s (1.44$\times$). Ours (sync), which keeps those fused operators and moves the frozen path to the NPU with synchronous requests, shortens it to 62.4\,s, a further 4.03$\times$ and 5.79$\times$ overall. Recorded clocks are similar across configurations, and the two rounds run in opposite orders from the same 42\,$^\circ$C start (Appendix~\ref{app:baselines}). Two findings explain the gain: after four basic CPU fixes, the frozen path still takes 55\% of self CPU time for one training example on P1 and about 40 of 68.6\,s on P2 (Figure~\ref{fig:motivation}a), and at the same forward shape, HMX reaches 4.65 TFLOPS on P1 and about 5.3 on P2 with FP16 weights, against 0.178 and 0.780 on the CPU.

\noindent\textbf{Fused CPU operators.}
Fusion shrinks the CPU residual that offload exposes. On P2, a zero-returning stub for the frozen path leaves 26.7\,s of self CPU time per training example, led by small operators (9.1\,s), casts and copies (6.2\,s), adapter products (4.8\,s), and GDN recurrence (4.3\,s). Operator tests reduce recurrent cost by about 5.8$\times$ on P1 and 3.8$\times$ on P2. Within a layer on a 744-token CPU microbenchmark, the fused LoRA operator reduces time from 9.4 to 2.2\,s. The fused vocabulary head takes 216\,ms for forward and backward at 38 answer positions, against 595\,ms for unfused slicing at 32 positions. Arithmetic, mask, threading, and cross-platform checks validate all fused operators.

\noindent\textbf{Overlap.}
Overlap hides adapter work behind NPU requests (Section~\ref{sec:overlap}) and adds to offload: in two cold-step trials in opposite orders, it cuts step time from 65.3 to 59.8\,s on P1 and from 60.2 to 52.4\,s on P2, with identical losses and no unconsumed requests. In a profile of a 744-token P2 example, overlap cuts the exposed backward-call time from 1.00\,s to 0.20\,s. In Table~\ref{tab:runs}, updates with overlap take 7.82 hours on P1 (8.51 without) and 8.57 on P2 (9.01 without). A six-thread P2 repeat that starts at full battery and stays at thermal status 0 takes 8.09 active hours (Appendix~\ref{app:energy}).

\noindent\textbf{After offload.}
To see what remains after offload, we profile a 744-token P2 example with overlap: client calls, including exposed NPU waiting, account for 15\% of self CPU time, against 56\% for fused CPU operators. The example takes 13.55\,s with profiling enabled and records 13.3\,s of self CPU time, of which client calls take 2.05\,s and fused recurrence, adapters, and elementwise operations take 7.43\,s. In this profile, the CPU--NPU boundary thus takes a small share of self CPU time, and the fused CPU work the largest. The input-gradient kernels take 0.51--5.60\,ms per projection at 840 tokens (Table~\ref{tab:kernels}); on V81, they reach 13.0 TFLOPS in FP16 at 840 tokens and 17.0 in Q8 at 2,048 tokens.

\begin{table}[t]
\caption{Input-gradient kernel latency at $M=840$ tokens.}
\label{tab:kernels}
\centering\tabstyle
\begin{tabular}{@{}lrr@{}}
\toprule
Projection & V79 (ms) & V81 (ms)\\
\midrule
FP16 GDN qkv & 5.60 & 5.00\\
FP16 GDN output gate & 2.19 & 2.28\\
FP16 GDN output & 1.63 & 1.35\\
Q8 MLP gate/up & 3.84 & 3.28\\
Q8 MLP down & 2.81 & 2.77\\
Q8 attention q & 3.21 & 2.42\\
Q8 attention k/v & 0.62 & 0.51\\
Q8 attention o & 1.46 & 1.19\\
\bottomrule
\end{tabular}
\vspace{-1.5em}
\end{table}

\noindent\textbf{Input-gradient kernels.}
Table~\ref{tab:kernels} lists kernel-only input-gradient latency for representative projections on both phones; each gate/up or k/v row is one product. FP16 rows come from the FP16 regression build and Q8 rows from the build used for training. The 13.0 TFLOPS on V81 come from the FP16 recurrent output projection at $M=840$, and the 17.0 TFLOPS from the Q8 attention output projection at $M=2048$. Fixed cost appears at small token counts: V79's Q8 MLP gate takes 0.80\,ms at $M=64$, 1.96\,ms at 429, and 8.22\,ms at 2,048. Operand exchange is cheap: in the V81 simulator, reusing the forward tile as the activation operand takes 76 cycles per tile, against 152--182 cycles for expanding and rearranging it as a weight operand.

\subsection{Numerical fidelity (RQ3)}\label{sec:fidelity}

RQ3 asks whether inference weights and FP16 outputs alter what the model learns. All ten checks in Table~\ref{tab:fidelity}, from kernels to task accuracy, stay near their references.

\noindent\textbf{Kernels.}
Input-gradient kernels meet the $10^{-6}$ NMSE target, and one-hot probes recover the forward effective weights bit for bit on both phones for tested shapes. Both phones pass all 260 startup checks for the full FP16 and Q8 configuration; the modified V79 build passes all 759 cases of ggml's forward matrix-multiply operation (\texttt{MUL\_MAT}) unchanged from the base build. The modified and base V79 builds keep the same 66 forward kernels, and alternating timing gives a geometric-mean new-to-base ratio of 1.007. The V79 support test passes all 55 supported cases and rejects all 30 unsupported ones, and the service's resident memory stays unchanged over 10,000 client calls.

\begin{table}[t]
\caption{Gradient error of frozen-weight precision policies.}
\label{tab:precision_full}
\centering\tabstyle
\begin{tabular}{@{}lrrr@{}}
\toprule
 & & \multicolumn{2}{c@{}}{Cosine}\\
\cmidrule(l){3-4}
Policy & Error (\%) & Global & Worst\\
\midrule
BF16 & 7.7 & 0.9971 & 0.93\\
FP16 & 9.3 & 0.9957 & 0.90\\
Q8 & 18.8 & 0.9822 & 0.47\\
Q8 + gates FP16 & 20.2 & 0.9795 & 0.37\\
Q8 + GDN FP16 (ours) & 9.6 & 0.9954 & 0.96\\
Q4 & 98.0 & 0.6400 & $-0.33$\\
\bottomrule
\end{tabular}
\vspace{-1em}
\end{table}

\noindent\textbf{Gradients.}
Our precision policy holds weight-format gradient error at the FP16 level (Table~\ref{tab:precision_full}): 9.6\% with the five recurrent projection types in FP16, against 9.3\% for all-FP16 weights, 18.8\% for uniform Q8, and 20.2\% with only the two gates in FP16. Under our policy, per-group errors of the recurrent projections fall to 7--14\%, from 24--42\% under uniform Q8, and the worst-module cosine similarity rises from 0.47 to 0.96. In an end-to-end integration harness, the emulated HMX path gives 10.15\% gradient error against FP32, below the 10.9\% of the BF16 training path and less than half the 21.1\% of a CPU Q8 path that quantizes activations. The fused GDN operator further lowers the BF16 path's error from 10.9\% to 8.8\%, and its 20 test cases agree bit for bit across x86 and Arm CPUs, thread counts, and replay. For the remaining FP16 overflow, the range guard flags no rows in complete hybrid Qwen3.5 updates, and the Q3-4B LaMP-2 run retries 52 flagged rows across the full epoch without service errors or restarts.

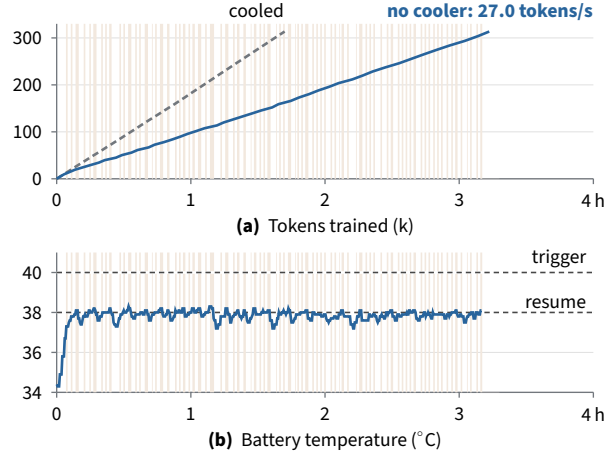
\begin{figure}[t]
\centering
\tikzset{
  chart base/.style={fill=papergray!20,draw=papergray,line width=0.5pt},
  chart base2/.style={fill=papergray!50,draw=papergray,line width=0.5pt},
  chart ours/.style={fill=paperblue,draw=paperblue,line width=0.5pt},
  chart warn/.style={fill=paperorange!50,draw=paperorange,line width=0.5pt},
  chart band/.style={fill=paperorange!15,draw=none},
  chart line ours/.style={draw=paperblue,line width=1pt},
  chart line base/.style={draw=papergray,line width=1pt,dash pattern=on 2.5pt off 1.5pt},
  chart line ref/.style={draw=uxink,line width=0.6pt,dash pattern=on 2pt off 1.5pt,
    preaction={draw=white,line width=1pt,solid}},
  chart val/.style={ux/value,anchor=west},
  chart val ours/.style={ux/value,anchor=west,text=paperblue,font=\uxfont\bfseries},
  chart title/.style={font=\uxfont,inner sep=0pt,anchor=base},
}
\newcommand{\ftitledrop}{20pt}
\newcommand{\ftitle}[3]{\node[chart title] at ([yshift=-\ftitledrop]#1.south)
  {\textbf{(#2)}\hspace{3pt}#3};}
\newcommand{\chartkey}[2]{{\setlength{\fboxsep}{0pt}\setlength{\fboxrule}{0.5pt}%
  \fcolorbox{#2}{#1}{\rule{0pt}{4.5pt}\rule{4.5pt}{0pt}}}}
\newcommand{\fgrid}[3]{\draw[/pgfplots/every axis grid]
  ({axis cs:#1,0} |- #2) -- ({axis cs:#1,0} |- #3);}
\newcommand{\thW}{7.1cm}
\newcommand{\thPanelA}[1][]{%
\begin{axis}[ux axis,name=ta,#1,width=\thW,height=2.05cm,
  xmin=0,xmax=4,ymin=0,ymax=330,xtick={0,1,2,3,4},xticklabels={0,1,2,3,4\,h},
  ytick={0,100,200,300},ymajorgrids=true]
  \addplot[chart band] coordinates {
    (0,0) (0.0692,0) (0.0692,330) (0.0817,330) (0.0817,0) (0.1054,0) (0.1054,330) (0.1174,330)
    (0.1174,0) (0.1407,0) (0.1407,330) (0.1645,330) (0.1645,0) (0.2014,0) (0.2014,330) (0.2136,330)
    (0.2136,0) (0.2432,0) (0.2432,330) (0.2550,330) (0.2550,0) (0.2728,0) (0.2728,330) (0.2968,330)
    (0.2968,0) (0.3321,0) (0.3321,330) (0.3441,330) (0.3441,0) (0.3617,0) (0.3617,330) (0.3736,330)
    (0.3736,0) (0.3913,0) (0.3913,330) (0.4091,330) (0.4091,0) (0.4677,0) (0.4677,330) (0.4796,330)
    (0.4796,0) (0.4972,0) (0.4972,330) (0.5091,330) (0.5091,0) (0.5325,0) (0.5325,330) (0.5563,330)
    (0.5563,0) (0.5854,0) (0.5854,330) (0.5972,330) (0.5972,0) (0.6149,0) (0.6149,330) (0.6268,330)
    (0.6268,0) (0.6564,0) (0.6564,330) (0.6806,330) (0.6806,0) (0.7100,0) (0.7100,330) (0.7160,330)
    (0.7160,0) (0.7339,0) (0.7339,330) (0.7459,330) (0.7459,0) (0.7697,0) (0.7697,330) (0.7934,330)
    (0.7934,0) (0.8227,0) (0.8227,330) (0.8406,330) (0.8406,0) (0.8822,0) (0.8822,330) (0.8942,330)
    (0.8942,0) (0.9118,0) (0.9118,330) (0.9236,330) (0.9236,0) (0.9413,0) (0.9413,330) (0.9589,330)
    (0.9589,0) (0.9824,0) (0.9824,330) (0.9943,330) (0.9943,0) (1.0118,0) (1.0118,330) (1.0296,330)
    (1.0296,0) (1.0531,0) (1.0531,330) (1.0771,330) (1.0771,0) (1.1066,0) (1.1066,330) (1.1185,330)
    (1.1185,0) (1.1425,0) (1.1425,330) (1.1729,330) (1.1729,0) (1.2204,0) (1.2204,330) (1.2322,330)
    (1.2322,0) (1.2559,0) (1.2559,330) (1.2795,330) (1.2795,0) (1.3147,0) (1.3147,330) (1.3266,330)
    (1.3266,0) (1.3442,0) (1.3442,330) (1.3622,330) (1.3622,0) (1.3918,0) (1.3918,330) (1.4099,330)
    (1.4099,0) (1.4461,0) (1.4461,330) (1.4521,330) (1.4521,0) (1.4700,0) (1.4700,330) (1.4881,330)
    (1.4881,0) (1.5117,0) (1.5117,330) (1.5235,330) (1.5235,0) (1.5413,0) (1.5413,330) (1.5529,330)
    (1.5529,0) (1.5704,0) (1.5704,330) (1.5943,330) (1.5943,0) (1.6352,0) (1.6352,330) (1.6411,330)
    (1.6411,0) (1.6585,0) (1.6585,330) (1.6703,330) (1.6703,0) (1.6881,0) (1.6881,330) (1.6999,330)
    (1.6999,0) (1.7236,0) (1.7236,330) (1.7471,330) (1.7471,0) (1.7764,0) (1.7764,330) (1.7821,330)
    (1.7821,0) (1.7938,0) (1.7938,330) (1.8057,330) (1.8057,0) (1.8232,0) (1.8232,330) (1.8349,330)
    (1.8349,0) (1.8524,0) (1.8524,330) (1.8757,330) (1.8757,0) (1.9052,0) (1.9052,330) (1.9170,330)
    (1.9170,0) (1.9352,0) (1.9352,330) (1.9468,330) (1.9468,0) (1.9646,0) (1.9646,330) (1.9763,330)
    (1.9763,0) (1.9939,0) (1.9939,330) (2.0059,330) (2.0059,0) (2.0413,0) (2.0413,330) (2.0529,330)
    (2.0529,0) (2.0706,0) (2.0706,330) (2.0824,330) (2.0824,0) (2.1060,0) (2.1060,330) (2.1118,330)
    (2.1118,0) (2.1235,0) (2.1235,330) (2.1411,330) (2.1411,0) (2.1649,0) (2.1649,330) (2.1770,330)
    (2.1770,0) (2.2307,0) (2.2307,330) (2.2367,330) (2.2367,0) (2.2545,0) (2.2545,330) (2.2784,330)
    (2.2784,0) (2.3018,0) (2.3018,330) (2.3135,330) (2.3135,0) (2.3310,0) (2.3310,330) (2.3428,330)
    (2.3428,0) (2.3604,0) (2.3604,330) (2.3722,330) (2.3722,0) (2.4132,0) (2.4132,330) (2.4250,330)
    (2.4250,0) (2.4602,0) (2.4602,330) (2.4721,330) (2.4721,0) (2.4896,0) (2.4896,330) (2.5075,330)
    (2.5075,0) (2.5427,0) (2.5427,330) (2.5543,330) (2.5543,0) (2.5718,0) (2.5718,330) (2.5835,330)
    (2.5835,0) (2.6007,0) (2.6007,330) (2.6127,330) (2.6127,0) (2.6303,0) (2.6303,330) (2.6421,330)
    (2.6421,0) (2.6597,0) (2.6597,330) (2.6777,330) (2.6777,0) (2.7129,0) (2.7129,330) (2.7246,330)
    (2.7246,0) (2.7421,0) (2.7421,330) (2.7538,330) (2.7538,0) (2.7713,0) (2.7713,330) (2.7832,330)
    (2.7832,0) (2.8066,0) (2.8066,330) (2.8182,330) (2.8182,0) (2.8359,0) (2.8359,330) (2.8477,330)
    (2.8477,0) (2.8650,0) (2.8650,330) (2.8768,330) (2.8768,0) (2.9060,0) (2.9060,330) (2.9177,330)
    (2.9177,0) (2.9410,0) (2.9410,330) (2.9527,330) (2.9527,0) (2.9702,0) (2.9702,330) (2.9877,330)
    (2.9877,0) (3.0110,0) (3.0110,330) (3.0227,330) (3.0227,0) (3.0578,0) (3.0578,330) (3.0638,330)
    (3.0638,0) (3.0813,0) (3.0813,330) (3.0931,330) (3.0931,0) (3.1282,0) (3.1282,330) (3.1399,330)
    (3.1399,0) (3.1572,0) (3.1572,330) (3.1717,330) (3.1717,0) (3.3,0)} -- cycle;
  \addplot[chart line base] coordinates {
    (0.000,0.0) (0.026,4.7) (0.052,9.4) (0.079,14.3) (0.106,19.3) (0.133,24.3) (0.161,29.4) (0.190,34.5)
    (0.219,39.7) (0.248,45.0) (0.278,50.4) (0.309,55.8) (0.340,61.3) (0.370,66.9) (0.402,72.5) (0.433,78.2)
    (0.465,83.9) (0.497,89.8) (0.529,95.7) (0.562,101.7) (0.595,107.7) (0.627,113.7) (0.661,119.9) (0.695,126.1)
    (0.731,132.5) (0.767,138.9) (0.803,145.5) (0.840,152.1) (0.878,158.9) (0.915,165.9) (0.954,173.1) (0.992,180.5)
    (1.032,188.1) (1.073,195.8) (1.115,203.7) (1.157,211.7) (1.200,219.9) (1.245,228.4) (1.292,237.2) (1.339,246.2)
    (1.388,255.3) (1.438,264.7) (1.488,274.1) (1.540,283.8) (1.592,293.6) (1.645,303.6) (1.699,313.6)};
  \addplot[chart line ours] coordinates {
    (0.000,0.0) (0.026,4.7) (0.054,9.4) (0.095,14.3) (0.136,19.3) (0.194,24.3) (0.254,29.4) (0.315,34.5)
    (0.358,39.7) (0.442,45.0) (0.486,50.4) (0.558,55.8) (0.603,61.3) (0.686,66.9) (0.729,72.5) (0.801,78.2)
    (0.863,83.9) (0.922,89.8) (0.976,95.7) (1.041,101.7) (1.103,107.7) (1.195,113.7) (1.244,119.9) (1.312,126.1)
    (1.382,132.5) (1.453,138.9) (1.524,145.5) (1.607,152.1) (1.653,158.9) (1.745,165.9) (1.806,173.1) (1.883,180.5)
    (1.950,188.1) (2.030,195.8) (2.101,203.7) (2.204,211.7) (2.281,219.9) (2.354,228.4) (2.449,237.2) (2.554,246.2)
    (2.642,255.3) (2.733,264.7) (2.827,274.1) (2.928,283.8) (3.041,293.6) (3.138,303.6) (3.224,313.6)};
  \node[ux/label,inner sep=0pt,anchor=base east] at ([yshift=2.5pt]axis cs:1.699,330) {cooled};
  \node[ux/label,inner sep=0pt,anchor=base east,text=paperblue,font=\uxfont\bfseries]
    at ([yshift=2.5pt]axis cs:4,330) {no cooler: 27.0 tokens/s};
\end{axis}}
\newcommand{\thPanelB}[1][]{%
\begin{axis}[ux axis,name=tb,#1,width=\thW,height=1.85cm,
  xmin=0,xmax=4,ymin=34,ymax=41,xtick={0,1,2,3,4},xticklabels={0,1,2,3,4\,h},
  ytick={34,36,38,40},ymajorgrids=true]
  \addplot[chart band] coordinates {
    (0,34) (0.0692,34) (0.0692,41) (0.0817,41) (0.0817,34) (0.1054,34) (0.1054,41) (0.1174,41)
    (0.1174,34) (0.1407,34) (0.1407,41) (0.1645,41) (0.1645,34) (0.2014,34) (0.2014,41) (0.2136,41)
    (0.2136,34) (0.2432,34) (0.2432,41) (0.2550,41) (0.2550,34) (0.2728,34) (0.2728,41) (0.2968,41)
    (0.2968,34) (0.3321,34) (0.3321,41) (0.3441,41) (0.3441,34) (0.3617,34) (0.3617,41) (0.3736,41)
    (0.3736,34) (0.3913,34) (0.3913,41) (0.4091,41) (0.4091,34) (0.4677,34) (0.4677,41) (0.4796,41)
    (0.4796,34) (0.4972,34) (0.4972,41) (0.5091,41) (0.5091,34) (0.5325,34) (0.5325,41) (0.5563,41)
    (0.5563,34) (0.5854,34) (0.5854,41) (0.5972,41) (0.5972,34) (0.6149,34) (0.6149,41) (0.6268,41)
    (0.6268,34) (0.6564,34) (0.6564,41) (0.6806,41) (0.6806,34) (0.7100,34) (0.7100,41) (0.7160,41)
    (0.7160,34) (0.7339,34) (0.7339,41) (0.7459,41) (0.7459,34) (0.7697,34) (0.7697,41) (0.7934,41)
    (0.7934,34) (0.8227,34) (0.8227,41) (0.8406,41) (0.8406,34) (0.8822,34) (0.8822,41) (0.8942,41)
    (0.8942,34) (0.9118,34) (0.9118,41) (0.9236,41) (0.9236,34) (0.9413,34) (0.9413,41) (0.9589,41)
    (0.9589,34) (0.9824,34) (0.9824,41) (0.9943,41) (0.9943,34) (1.0118,34) (1.0118,41) (1.0296,41)
    (1.0296,34) (1.0531,34) (1.0531,41) (1.0771,41) (1.0771,34) (1.1066,34) (1.1066,41) (1.1185,41)
    (1.1185,34) (1.1425,34) (1.1425,41) (1.1729,41) (1.1729,34) (1.2204,34) (1.2204,41) (1.2322,41)
    (1.2322,34) (1.2559,34) (1.2559,41) (1.2795,41) (1.2795,34) (1.3147,34) (1.3147,41) (1.3266,41)
    (1.3266,34) (1.3442,34) (1.3442,41) (1.3622,41) (1.3622,34) (1.3918,34) (1.3918,41) (1.4099,41)
    (1.4099,34) (1.4461,34) (1.4461,41) (1.4521,41) (1.4521,34) (1.4700,34) (1.4700,41) (1.4881,41)
    (1.4881,34) (1.5117,34) (1.5117,41) (1.5235,41) (1.5235,34) (1.5413,34) (1.5413,41) (1.5529,41)
    (1.5529,34) (1.5704,34) (1.5704,41) (1.5943,41) (1.5943,34) (1.6352,34) (1.6352,41) (1.6411,41)
    (1.6411,34) (1.6585,34) (1.6585,41) (1.6703,41) (1.6703,34) (1.6881,34) (1.6881,41) (1.6999,41)
    (1.6999,34) (1.7236,34) (1.7236,41) (1.7471,41) (1.7471,34) (1.7764,34) (1.7764,41) (1.7821,41)
    (1.7821,34) (1.7938,34) (1.7938,41) (1.8057,41) (1.8057,34) (1.8232,34) (1.8232,41) (1.8349,41)
    (1.8349,34) (1.8524,34) (1.8524,41) (1.8757,41) (1.8757,34) (1.9052,34) (1.9052,41) (1.9170,41)
    (1.9170,34) (1.9352,34) (1.9352,41) (1.9468,41) (1.9468,34) (1.9646,34) (1.9646,41) (1.9763,41)
    (1.9763,34) (1.9939,34) (1.9939,41) (2.0059,41) (2.0059,34) (2.0413,34) (2.0413,41) (2.0529,41)
    (2.0529,34) (2.0706,34) (2.0706,41) (2.0824,41) (2.0824,34) (2.1060,34) (2.1060,41) (2.1118,41)
    (2.1118,34) (2.1235,34) (2.1235,41) (2.1411,41) (2.1411,34) (2.1649,34) (2.1649,41) (2.1770,41)
    (2.1770,34) (2.2307,34) (2.2307,41) (2.2367,41) (2.2367,34) (2.2545,34) (2.2545,41) (2.2784,41)
    (2.2784,34) (2.3018,34) (2.3018,41) (2.3135,41) (2.3135,34) (2.3310,34) (2.3310,41) (2.3428,41)
    (2.3428,34) (2.3604,34) (2.3604,41) (2.3722,41) (2.3722,34) (2.4132,34) (2.4132,41) (2.4250,41)
    (2.4250,34) (2.4602,34) (2.4602,41) (2.4721,41) (2.4721,34) (2.4896,34) (2.4896,41) (2.5075,41)
    (2.5075,34) (2.5427,34) (2.5427,41) (2.5543,41) (2.5543,34) (2.5718,34) (2.5718,41) (2.5835,41)
    (2.5835,34) (2.6007,34) (2.6007,41) (2.6127,41) (2.6127,34) (2.6303,34) (2.6303,41) (2.6421,41)
    (2.6421,34) (2.6597,34) (2.6597,41) (2.6777,41) (2.6777,34) (2.7129,34) (2.7129,41) (2.7246,41)
    (2.7246,34) (2.7421,34) (2.7421,41) (2.7538,41) (2.7538,34) (2.7713,34) (2.7713,41) (2.7832,41)
    (2.7832,34) (2.8066,34) (2.8066,41) (2.8182,41) (2.8182,34) (2.8359,34) (2.8359,41) (2.8477,41)
    (2.8477,34) (2.8650,34) (2.8650,41) (2.8768,41) (2.8768,34) (2.9060,34) (2.9060,41) (2.9177,41)
    (2.9177,34) (2.9410,34) (2.9410,41) (2.9527,41) (2.9527,34) (2.9702,34) (2.9702,41) (2.9877,41)
    (2.9877,34) (3.0110,34) (3.0110,41) (3.0227,41) (3.0227,34) (3.0578,34) (3.0578,41) (3.0638,41)
    (3.0638,34) (3.0813,34) (3.0813,41) (3.0931,41) (3.0931,34) (3.1282,34) (3.1282,41) (3.1399,41)
    (3.1399,34) (3.1572,34) (3.1572,41) (3.1717,41) (3.1717,34) (3.3,34)} -- cycle;
  \draw[chart line ref] (axis cs:0,40) -- (axis cs:4,40);
  \draw[chart line ref] (axis cs:0,38) -- (axis cs:4,38);
  \addplot[chart line ours] coordinates {
    (0.001,34.4) (0.007,34.3) (0.013,34.3) (0.018,34.3) (0.024,34.9) (0.030,34.9) (0.036,34.9) (0.042,35.8) (0.048,35.8)
    (0.054,35.8) (0.060,36.7) (0.066,36.7) (0.072,37.3) (0.079,37.3) (0.085,37.3) (0.091,37.5) (0.097,37.6) (0.103,37.6)
    (0.108,37.8) (0.114,37.8) (0.120,37.8) (0.126,37.8) (0.132,37.9) (0.138,37.9) (0.144,38.1) (0.149,38.1) (0.156,38.1)
    (0.161,37.7) (0.168,37.7) (0.173,37.5) (0.179,37.4) (0.186,37.4) (0.192,37.6) (0.198,37.8) (0.205,37.8) (0.211,37.9)
    (0.217,37.8) (0.223,37.8) (0.229,37.8) (0.234,37.9) (0.240,38.1) (0.246,38.1) (0.252,38.0) (0.258,38.0) (0.264,38.0)
    (0.270,38.1) (0.276,38.2) (0.282,38.2) (0.288,38.1) (0.294,37.9) (0.300,37.9) (0.306,37.6) (0.312,37.6) (0.318,37.6)
    (0.323,37.8) (0.329,37.8) (0.335,37.8) (0.341,38.0) (0.347,38.0) (0.353,38.0) (0.359,38.0) (0.365,38.0) (0.371,38.0)
    (0.377,38.0) (0.383,38.0) (0.388,38.0) (0.394,38.2) (0.400,38.2) (0.406,38.0) (0.412,37.9) (0.418,37.9) (0.424,37.5)
    (0.430,37.4) (0.436,37.4) (0.441,37.3) (0.447,37.3) (0.453,37.3) (0.459,37.5) (0.465,37.7) (0.471,37.7) (0.477,37.9)
    (0.483,37.9) (0.488,37.9) (0.494,38.0) (0.500,38.1) (0.506,38.0) (0.512,38.0) (0.518,38.0) (0.524,38.1) (0.529,38.1)
    (0.536,38.3) (0.541,38.2) (0.547,38.2) (0.553,38.0) (0.559,37.8) (0.565,37.8) (0.571,37.7) (0.577,37.7) (0.583,37.7)
    (0.588,37.9) (0.594,37.9) (0.600,37.9) (0.606,37.9) (0.612,37.9) (0.618,37.9) (0.624,38.0) (0.630,38.0) (0.636,38.0)
    (0.642,37.9) (0.648,37.9) (0.653,37.9) (0.659,38.1) (0.666,38.1) (0.672,38.1) (0.678,37.9) (0.684,37.9) (0.689,37.6)
    (0.695,37.6) (0.701,37.6) (0.707,37.8) (0.713,37.9) (0.719,37.9) (0.725,37.9) (0.731,38.0) (0.737,38.0) (0.743,38.0)
    (0.749,38.0) (0.755,38.0) (0.761,38.1) (0.767,38.1) (0.773,38.2) (0.779,38.2) (0.784,38.2) (0.790,37.9) (0.796,37.8)
    (0.802,37.8) (0.808,37.8) (0.814,37.9) (0.820,37.9) (0.826,38.1) (0.832,38.1) (0.838,38.0) (0.844,37.9) (0.849,37.6)
    (0.856,37.6) (0.861,37.6) (0.868,37.7) (0.873,37.8) (0.879,37.8) (0.885,38.1) (0.891,38.0) (0.897,38.0) (0.903,38.0)
    (0.909,38.1) (0.915,38.1) (0.921,38.0) (0.927,38.0) (0.933,38.0) (0.938,38.2) (0.944,38.2) (0.950,38.2) (0.956,37.8)
    (0.962,37.8) (0.968,37.8) (0.974,37.9) (0.979,37.9) (0.985,37.9) (0.991,38.0) (0.997,38.0) (1.003,38.0) (1.009,38.2)
    (1.015,38.2) (1.021,38.2) (1.027,38.0) (1.033,38.0) (1.038,38.0) (1.044,38.1) (1.050,38.1) (1.056,38.2) (1.062,38.2)
    (1.068,38.2) (1.074,38.0) (1.080,37.8) (1.086,37.8) (1.092,37.8) (1.098,37.9) (1.104,37.9) (1.109,38.0) (1.116,38.0)
    (1.121,38.0) (1.127,38.0) (1.133,38.1) (1.139,38.3) (1.146,38.3) (1.152,38.3) (1.158,38.2) (1.164,38.2) (1.170,37.7)
    (1.176,37.6) (1.182,37.6) (1.188,37.2) (1.194,37.2) (1.200,37.2) (1.206,37.4) (1.212,37.4) (1.218,37.4) (1.223,37.8)
    (1.229,37.8) (1.235,37.8) (1.241,37.8) (1.247,37.8) (1.253,37.8) (1.259,38.2) (1.265,38.2) (1.271,38.2) (1.276,37.9)
    (1.283,37.9) (1.288,37.9) (1.294,37.5) (1.300,37.5) (1.306,37.7) (1.312,37.8) (1.318,37.8) (1.324,37.9) (1.329,37.9)
    (1.335,37.9) (1.341,38.0) (1.347,38.1) (1.353,38.1) (1.359,37.9) (1.365,37.8) (1.371,37.8) (1.377,37.7) (1.383,37.8)
    (1.389,37.8) (1.395,38.1) (1.401,38.1) (1.407,37.9) (1.413,37.9) (1.419,37.6) (1.425,37.5) (1.431,37.5) (1.437,37.5)
    (1.443,37.7) (1.449,37.7) (1.455,37.8) (1.461,37.9) (1.467,38.2) (1.473,38.2) (1.479,38.2) (1.485,37.9) (1.491,37.9)
    (1.497,37.9) (1.503,37.9) (1.509,37.9) (1.515,37.9) (1.521,37.9) (1.526,38.0) (1.533,38.0) (1.538,38.1) (1.544,38.0)
    (1.550,38.0) (1.556,38.0) (1.562,38.0) (1.568,38.0) (1.573,38.1) (1.579,38.1) (1.586,38.1) (1.591,37.6) (1.597,37.6)
    (1.603,37.6) (1.609,37.2) (1.615,37.2) (1.621,37.2) (1.626,37.4) (1.632,37.4) (1.638,37.4) (1.644,37.7) (1.650,37.7)
    (1.656,38.0) (1.661,37.9) (1.667,37.9) (1.673,37.9) (1.679,38.0) (1.685,38.0) (1.691,38.1) (1.697,38.0) (1.703,38.0)
    (1.709,37.9) (1.715,38.0) (1.721,38.0) (1.727,38.2) (1.733,38.1) (1.738,38.1) (1.744,37.8) (1.750,37.5) (1.756,37.5)
    (1.762,37.5) (1.768,37.6) (1.774,37.7) (1.779,37.7) (1.785,37.9) (1.791,37.9) (1.797,37.9) (1.803,37.9) (1.809,37.9)
    (1.814,37.9) (1.820,38.0) (1.826,38.0) (1.832,38.0) (1.838,38.0) (1.844,38.0) (1.849,38.0) (1.855,38.1) (1.861,38.1)
    (1.867,38.1) (1.873,37.7) (1.879,37.7) (1.884,37.7) (1.890,37.6) (1.896,37.6) (1.902,37.6) (1.908,38.0) (1.914,38.0)
    (1.920,38.0) (1.926,37.9) (1.932,37.9) (1.938,37.9) (1.944,37.9) (1.950,37.9) (1.956,37.9) (1.962,37.9) (1.968,37.9)
    (1.973,37.8) (1.979,37.8) (1.985,37.8) (1.991,37.9) (1.997,37.9) (2.003,37.9) (2.009,37.7) (2.015,37.6) (2.021,37.6)
    (2.027,37.5) (2.033,37.6) (2.038,37.6) (2.044,37.7) (2.050,37.7) (2.056,37.7) (2.062,37.7) (2.068,37.8) (2.074,37.8)
    (2.079,37.8) (2.085,37.7) (2.091,37.7) (2.097,37.7) (2.103,37.9) (2.109,37.9) (2.115,37.9) (2.121,38.1) (2.126,38.1)
    (2.132,38.1) (2.138,37.9) (2.144,37.8) (2.150,37.8) (2.156,37.8) (2.162,37.8) (2.168,37.8) (2.174,37.8) (2.180,37.8)
    (2.186,37.8) (2.192,37.4) (2.198,37.4) (2.204,37.4) (2.210,37.2) (2.216,37.2) (2.222,37.5) (2.228,37.5) (2.234,37.5)
    (2.240,37.8) (2.246,37.8) (2.251,37.8) (2.258,38.1) (2.263,38.1) (2.269,38.1) (2.275,37.6) (2.281,37.6) (2.287,37.6)
    (2.293,37.6) (2.299,37.6) (2.305,37.6) (2.311,37.6) (2.316,37.6) (2.322,37.6) (2.328,37.9) (2.334,37.9) (2.340,37.8)
    (2.346,37.8) (2.352,37.8) (2.358,37.9) (2.363,37.9) (2.369,37.9) (2.375,37.8) (2.381,37.8) (2.387,37.6) (2.393,37.5)
    (2.399,37.6) (2.404,37.6) (2.410,37.8) (2.416,37.8) (2.422,37.9) (2.428,37.8) (2.434,37.8) (2.440,37.7) (2.446,37.7)
    (2.451,37.8) (2.457,38.0) (2.463,38.0) (2.469,37.9) (2.475,38.0) (2.481,38.0) (2.487,38.1) (2.493,38.1) (2.499,38.1)
    (2.504,37.8) (2.511,37.8) (2.516,37.8) (2.522,37.6) (2.528,37.6) (2.534,37.6) (2.540,37.9) (2.546,37.9) (2.551,37.9)
    (2.557,37.8) (2.563,37.8) (2.569,37.8) (2.575,37.9) (2.581,37.9) (2.586,37.9) (2.592,37.9) (2.598,37.9) (2.604,37.9)
    (2.610,38.0) (2.616,37.9) (2.621,37.9) (2.627,38.1) (2.633,38.1) (2.639,38.0) (2.645,38.0) (2.651,38.0) (2.657,38.1)
    (2.663,38.1) (2.669,38.1) (2.675,38.0) (2.681,37.7) (2.686,37.7) (2.693,37.6) (2.698,37.6) (2.704,37.6) (2.710,37.8)
    (2.716,37.8) (2.722,37.8) (2.728,37.8) (2.733,37.9) (2.739,38.0) (2.745,38.0) (2.751,37.8) (2.757,37.9) (2.763,37.9)
    (2.768,38.0) (2.774,37.9) (2.780,37.9) (2.786,37.8) (2.792,37.7) (2.798,37.7) (2.804,38.0) (2.809,37.9) (2.815,37.9)
    (2.821,37.9) (2.827,37.9) (2.833,37.9) (2.839,37.9) (2.845,37.9) (2.851,37.9) (2.856,37.9) (2.862,37.9) (2.868,37.9)
    (2.874,37.9) (2.880,37.9) (2.886,37.9) (2.891,37.7) (2.897,37.7) (2.903,37.7) (2.909,37.9) (2.915,37.9) (2.921,37.9)
    (2.926,37.8) (2.932,37.9) (2.938,37.9) (2.944,38.0) (2.950,38.0) (2.956,38.0) (2.961,38.0) (2.967,38.0) (2.973,38.1)
    (2.979,38.1) (2.985,37.9) (2.991,37.8) (2.996,37.8) (3.002,37.8) (3.008,37.9) (3.014,37.9) (3.020,37.8) (3.026,37.8)
    (3.031,37.8) (3.037,37.5) (3.043,37.5) (3.049,37.5) (3.055,37.7) (3.061,37.8) (3.067,37.8) (3.073,37.9) (3.078,37.9)
    (3.084,37.9) (3.090,38.0) (3.096,38.0) (3.102,38.0) (3.108,37.7) (3.114,37.7) (3.119,37.7) (3.125,38.0) (3.131,38.0)
    (3.137,38.0) (3.143,37.9) (3.149,37.9) (3.154,37.9) (3.160,38.1) (3.166,38.1)};
  \node[ux/label,inner sep=0pt,anchor=base east] at ([xshift=-2.5pt,yshift=3.4pt]axis cs:4,40) {trigger};
  \node[ux/label,inner sep=0pt,anchor=base east] at ([xshift=-2.5pt,yshift=1.8pt]axis cs:4,38) {resume};
\end{axis}}
\begin{tikzpicture}
\thPanelB[at={(0,0)},anchor=south west]
\ftitle{tb}{b}{Battery temperature ($^\circ$C)}
\thPanelA[at={(0,1.85cm+6pt+\ftitledrop+1.75pt)},anchor=south west]
\ftitle{ta}{a}{Tokens trained (k)}
\end{tikzpicture}
\caption{Thermal pacing keeps training going on P2 with six pinned threads, the screen off, and no cooler. (a) Tokens trained over wall-clock time, against the cooled six-thread run on the same steps; (b) battery temperature, at most 38.3\,$^\circ$C and below the 40\,$^\circ$C trigger; Android thermal status 2 starts all 81 pauses (shaded).}
\label{fig:thermal}
\vspace{-1.5em}
\end{figure}

\noindent\textbf{Trajectories and tasks.}
Over 222 nightly-update steps, the phone's mean per-step loss gap to the GPU is smaller than the gap between GPU dropout seeds (Table~\ref{tab:fidelity}), and the phone falls within the four-seed range on 65.8\% of steps, close to the 60\% expected for an exchangeable fifth run.

\begin{table}[t]
\caption{LaMP-2 step-loss differences over 478 steps.}
\label{tab:lamp_traj}
\centering\tabstyle
\begin{tabular}{@{}lrrrr@{}}
\toprule
 & \multicolumn{2}{c}{Ours vs} & GPU 0 & In seed\\
\cmidrule(lr){2-3}
Model & GPU 0 & GPU 1 & vs GPU 1 & range (\%)\\
\midrule
Q3.5-2B & 0.0132 & 0.0121 & 0.0120 & 34.5\\
Q3-4B & 0.0206 & 0.0181 & 0.0158 & 25.9\\
L3.2-3B & 0.0106 & 0.0123 & 0.0112 & 31.6\\
\bottomrule
\end{tabular}
\vspace{-1.5em}
\end{table}

On LaMP-2, the pattern holds across all three models (Table~\ref{tab:lamp_traj}): the phone's mean step-loss gap to each GPU seed is comparable to the gap between the two seeds, and the phone falls within the seed range on 26--35\% of steps, near the 33\% expected for an exchangeable third run. A 30-step cross-phone L3.2-3B run further confirms that P2's per-step loss gap to P1 stays within the variation between GPU seeds.

\subsection{Energy and memory (RQ4)}\label{sec:cost}

Offload cuts energy per token by more than two thirds relative to C3. On the same 3,431-token batch on battery power, our system with overlap uses 0.2275\,J per token above idle, 32.2\% of C3's energy per token and 25.3\% of C0's (Figure~\ref{fig:headline}c). It finishes the batch soon enough that energy per token falls although its whole-device power draw rises to 13.3--13.8\,W, from about 11.0\,W for C3 (Appendix~\ref{app:energy}).

The precision policy also shrinks the resident model: the service holds 4.74\,GB of Q3.5-4B weights (2.02\,GB FP16 recurrent, 2.72\,GB Q8), against 7.14\,GB in all-FP16. In the six updates, the trainer peaks at 4.25--4.56\,GB, minimum available memory is 1.80\,GB on P1 and 1.74\,GB on P2, and the first optimizer step starts 45--77\,s after launch, once the NPU service loads its weights and validates its shapes.

\begin{table}[t]
\caption{First ten steps of three Qwen3.5 models on P1.}
\label{tab:sizes}
\centering\tabstyle
\begin{tabular}{@{}lrrr@{}}
\toprule
Size & Time (s) & Rate (tok/s) & RSS (GB)\\
\midrule
0.8B & 259 & 152.6 & 2.05\\
2B & 376 & 105.2 & 3.12\\
4B & 815 & 48.5 & 3.82\\
\bottomrule
\end{tabular}
\vspace{-2em}
\end{table}

Time and memory scale down with model size. Table~\ref{tab:sizes} reports the first ten steps of three Qwen3.5 sizes on P1 under the same cooling and screen conditions; RSS is the trainer's peak resident set size, excluding service weights. Maximum step-loss differences from the mean of two GPU runs are 0.0142, 0.0107, and 0.0049 for 0.8B, 2B, and 4B, against GPU-seed differences of 0.0076, 0.0107, and 0.0014. In 0.8B and 2B, the 16-output gates run in the CPU implementation with complete forward and backward. The resident service also loads Qwen3.5-9B: on P1, it holds all 248 projections (8.87\,GB) and passes 232 startup checks, and the phone's available memory falls from 10.2\,GB to 1.5\,GB.

\subsection{Training as a background job (RQ5)}\label{sec:sustained}

Training can also run without a cooler. With passive cooling only, a 339,380-token update on P2 completes in 2.64 hours with synchronous requests. The full nightly update needs pacing: under fixed load without a cooler, it drives P2 to Android thermal status 3 (severe) after about 18 minutes, and thermal pacing turns this into steady progress (Figure~\ref{fig:thermal}).

With thermal pacing on P2, six pinned threads, and no cooler, the paced run trains 313,592 tokens over 46 steps in 3.22 wall-clock hours and averages 27.0 tokens/s at a 64\% duty cycle, 53\% of the cooled six-thread run's rate on the same steps. Thermal status 2 starts all 81 pauses, and the battery stays at or below 38.3\,$^\circ$C and the CPU at or below 72.2\,$^\circ$C. Pausing and resuming do not change the computation: all 58 committed losses, including those of the 12 fixed-load steps before pacing starts, match the cooled runs bit for bit.

The job also yields and recovers: a P1 update that yields twice finishes in 7.82 hours with the same adapter as an uninterrupted repeat (7.87 hours). In service-kill tests on both phones, the training job restores the last commit and reproduces the subsequent losses and the final adapter; on P2, the restarted service listens again 54\,s after the failure.

\subsection{Other on-device training routes (RQ6)}\label{sec:alt}

\begin{table}[t]
\caption{Phone-GPU and NPU training rates on L3.2-3B.}
\label{tab:alt}
\centering\tabstyle
\begin{tabular}{@{}lrrr@{}}
\toprule
Phone & Phone GPU (tok/s) & Ours (tok/s) & Ratio\\
\midrule
P1 & 3.9 & 72.9 & 18.7$\times$\\
P2 & 5.1 & 65.1 & 12.8$\times$\\
\bottomrule
\end{tabular}
\vspace{-2em}
\end{table}

\noindent\textbf{Phone GPU.}
Fabric runs LoRA training on the GPUs of the same two phones, so we compare it with our path on a shared model and task. Both paths train L3.2-3B with rank 16 on seven projection types and LaMP-2 records. Fabric also converts weights to Q8, skips 9 longer records, and runs 128-position microbatches; our path uses the native template, variable-length records, and gradient accumulation. Because Fabric pads records to 512 positions under a ChatML answer mask, we compare useful tokens: we divide the native-template mean of 343 tokens per record by its time per padded record. On P2, Fabric runs for 20 minutes without device loss, over which we measure its rate.

On L3.2-3B, our system trains an order of magnitude faster, at 72.9 tokens/s on P1 over a full epoch and 65.1 on P2 over 30 steps, 18.7$\times$ and 12.8$\times$ Fabric's useful-token rate (Table~\ref{tab:alt}). On P1, all four Fabric attempts lose the Vulkan device, the longest after 6\,min 11\,s, and on both phones Fabric loses it on hybrid Q3.5-2B records of 128 or 512 positions (Appendix~\ref{app:baselines}). As a build check, we run Tether's published Qwen3-1.7B configuration: 13.3\,s per 128-position step on P1 and 10.1\,s on P2, against the reported 21\,s per step on an Adreno 830 GPU, P1's GPU~\cite{qvacbench}.

\noindent\textbf{Forward-only training.}
Backpropagation yields far more accurate adapters than MeZO's forward-only updates. With the same LoRA targets and data and a learning-rate sweep, MeZO reaches at most 37.86\% on Q3.5-2B (5,000 steps), 36.99\% on Q3-4B, and 34.83\% on \mbox{L3.2-3B}. One epoch of backpropagation on the phone reaches 58.96\%, 57.51\%, and 61.13\% on the three models, and no L3.2-3B MeZO checkpoint differs significantly from its base ($p\ge0.11$).

\section{Related Work}

Forward-only methods train through inference engines by estimating gradients from forward passes~\cite{mezo}, also on phones~\cite{mobizo}; we keep backpropagation. First-order trainers run it on edge and phone GPUs~\cite{pockengine,qvac}, and smaller networks train on phone DSPs and NPUs~\cite{mandheling,hexatrain}. We add the input-gradient product to the NPU's inference engine, so the frozen path of billion-parameter models backpropagates on the phone's NPU.

We build on LoRA~\cite{lora} and activation checkpointing~\cite{checkpoint}. Unlike weight quantization for output quality~\cite{gptq,awq}, we set each projection's precision by gradient sensitivity, and unlike a global loss scale~\cite{mixedprecision}, our range guard scales each row. Appendix~\ref{app:related} details each route.

\section{Conclusion}

An on-device LLM should learn from its user's data where that data lives. The phone's inference NPU can carry the expensive part of first-order LoRA training: the frozen path in both directions, computed from its resident inference weights. Around this offload, our system adds a gradient-sensitive precision policy, a per-row range guard, and a resumable runtime that fuses residual CPU work and overlaps requests. It completes nightly personalization updates of a 4B model on P1 and P2 with an external cooler and sustains thermally paced training on P2 without one. Its LaMP-2 adapters land within 1.01 points of GPU-trained ones on all three models, and on P1 it runs a controlled step 4.03$\times$ faster than an optimized CPU trainer. Because training asks the NPU only for products over the weights its inference engine already holds, the weights that serve an on-device LLM's inference also serve its fine-tuning.

\clearpage
\nobalance
\bibliographystyle{mlsys2025}
\bibliography{references}

\clearpage
\fancyhead[C]{\small\bfseries First-Order LoRA Training with Phone NPUs: Appendix}
\appendix
\numberwithin{table}{section}
\numberwithin{figure}{section}
\section{Numerical and Kernel Validation}\label{app:numerics}

\subsection{Precision policy measurements}

The precision study of Section~\ref{sec:precisionobs} compares concatenated LoRA gradients $g$ of Q3.5-4B for four evaluation-mode examples with an FP32 reference. It emulates block quantization, row scaling, FP16 operands and outputs, and FP32 accumulation. Table~\ref{tab:precision_full} in Section~\ref{sec:fidelity} reports the relative error $\|g-g_{\rm FP32}\|_2/\|g_{\rm FP32}\|_2$, the global cosine similarity, and the worst (minimum) cosine over modules. Its BF16 row stores frozen weights in BF16 without autocast, and its Q4 row quantizes them to 4-bit blocks with one scale each.

\subsection{Operand exchange}

Both NPUs have an 8\,MB VTCM, and HMX accumulates in 37-bit cells~\cite{hmx}. Q8 stores blocks of 32 signed 8-bit values with one FP16 scale, and device weights occupy $32\times32$ tiles of 1,088 bytes, including 64 scale bytes. One-hot probes recover the forward effective weights bit for bit on both phones for tested shapes.

\subsection{Nonfinite inputs and row retries}

HMX can return finite outputs for input rows that contain NaN. The CPU adapter branch receives the same input and propagates nonfinite values through the integrated layer. Integrated checks exercise nonfinite inputs and row retries.

\section{Task-Quality Details}\label{app:quality}

\subsection{LaMP-2}

Table~\ref{tab:lamp_f1} gives the macro-F1 of our phone-trained adapters and the two GPU-trained seeds on all 692 records (Section~\ref{sec:e2e}). Table~\ref{tab:lamp_full} lists each model's best MeZO checkpoint across sweeps. On Q3.5-2B, MeZO reaches 32.51--35.26\% after 500 steps, and the low-learning-rate run reaches 37.86\% at 5,000 steps and 37.72\% at 10,000 steps. Q3-4B MeZO checkpoints reach 34.54--36.99\%, and \mbox{L3.2-3B} MeZO checkpoints at 500--5,000 steps reach 30.49--34.83\%, none of them significantly different from its base ($p\ge0.11$).

\begin{table}[htb]
\caption{LaMP-2 development macro-F1 of the adapters.}
\label{tab:lamp_f1}
\centering\tabstyle
\begin{tabular}{@{}lrrr@{}}
\toprule
Model & Ours & GPU 0 & GPU 1\\
\midrule
Q3.5-2B & 0.5326 & 0.5309 & 0.5125\\
Q3-4B & 0.5128 & 0.5191 & 0.5187\\
L3.2-3B & 0.5614 & 0.5613 & 0.5564\\
\bottomrule
\end{tabular}
\vspace{-2em}
\end{table}

\subsection{Nightly personalization update}

On the 510-item held-out set, the phone adapter and the same-recipe GPU reference answer 241 and 243 correctly.

\section{Baselines}\label{app:baselines}

\subsection{Controlled CPU-to-NPU ladder}

Table~\ref{tab:ladder} gives the values plotted in Figure~\ref{fig:headline}b for C0--C3 (Section~\ref{sec:setup}) and Ours (sync) (Section~\ref{sec:speed}). All five configurations process the same 3,431-token batch on P1 with an external cooler, the screen off, and eight threads, in two rounds of opposite order, each starting at 42\,$^\circ$C.

\begin{table}[htb]
\caption{Controlled CPU-to-NPU step-time ladder on P1.\protect\vphantom{g}}
\label{tab:ladder}
\centering\tabstyle
\begin{tabular}{@{}lrr@{}}
\toprule
Configuration & Step (s) & Speedup vs C0\\
\midrule
C0 & 361.0 & 1.00$\times$\\
C1 + GDN & 310.9 & 1.16$\times$\\
C2 + LoRA & 271.8 & 1.33$\times$\\
C3 + head & 251.4 & 1.44$\times$\\
Ours (sync) & 62.4 & 5.79$\times$\\
\bottomrule
\end{tabular}
\vspace{-2em}
\end{table}

\subsection{Phone-GPU trainer}

All four P1 Fabric attempts lose the Vulkan device: two at the first step, at 512 and 64 positions, and two at 128 positions, whose last progress is at 6\,min 11\,s and 5\,min 29\,s. The 64-position failure occurs at a phone-GPU temperature of 54\,$^\circ$C. On hybrid Q3.5-2B, Fabric loses the device with 128- and 512-position records on both phones. Plain-text windows of 32 or 64 tokens run at 5.0--6.8 tokens/s on P1 and 6.5--7.5 on P2, and recurrent-backward checks pass 28 of 30 cases on P1 and 30 of 30 on P2 against the CPU.

\section{Energy and Sustained Operation}\label{app:energy}

\subsection{Energy measurement}

We estimate device energy from charge-counter brackets and measured voltage and subtract the measured idle draw of 108\,mA. On P2, the order Ours (with overlap), C3, C0, C0, C3, Ours gives two segments per configuration; each Ours segment contains four steps and each CPU segment one. The battery falls from 100\% to 84\%, and repeated segment losses are bit-identical. Table~\ref{tab:energy} lists the results; under load, Ours draws 13.3--13.8\,W and C3 about 11.0\,W.

\begin{table}[htb]
\caption{Whole-device energy per token above idle on P2.}
\label{tab:energy}
\centering\tabstyle
\begin{tabular}{@{}lrr@{}}
\toprule
Path & Energy (J/token) & Range (J/token)\\
\midrule
Ours & 0.2275 & 0.2234--0.2316\\
C3 & 0.7057 & 0.6950--0.7165\\
C0 & 0.9002 & 0.8853--0.9151\\
\bottomrule
\end{tabular}
\vspace{-2em}
\end{table}

\subsection{Operating state of complete updates}

P2's eight-thread overlap run spends 99\% of its samples at thermal status 2, with a median CPU temperature of 100.4\,$^\circ$C; the six-thread run starts at full battery and stays at status 0. The screen wakes during parts of the two P1 overlap runs (27\% and 44\% of state samples). Minimum available memory is 1.80\,GB on P1 and 1.74\,GB on P2, and maximum swap is 2.87 and 2.95\,GB, respectively.

\section{Detailed Related Work}\label{app:related}

\paragraph{Parameter-efficient tuning.}
LoRA~\cite{lora} and adapter layers~\cite{houlsby} shrink the trainable state. QLoRA~\cite{qlora} keeps the frozen base in low precision while computing in higher precision, and MeSP restructures adapter backward to save memory while keeping gradients mathematically identical~\cite{mesp}. They reduce what training stores; we execute what remains, the frozen path, on a phone's inference NPU.

\paragraph{Updates through inference engines.}
FwdLLM fine-tunes LLMs for federated learning with perturbed inferences~\cite{fwdllm}, MobiZO and MobiEdit run forward-only updates through mobile inference engines~\cite{mobizo,mobiedit}, and MeZO estimates gradients from forward passes to save optimizer memory~\cite{mezo}. These methods avoid backward kernels; we instead add the input-gradient product to the NPU's inference engine, so training on the phone keeps backpropagated gradients.

\paragraph{Training on mobile and edge hardware.}
On phones, offloading prefill to the NPU already speeds up LLM inference~\cite{llmnpu}. TinyTL freezes weights and learns small modules to cut activation memory~\cite{tinytl}, and training under 256\,KB brings learning to microcontrollers~\cite{tinytraining}. Mandheling trains deep neural networks with mixed precision across a phone's CPU and DSP~\cite{mandheling}. PockEngine compiles sparse backpropagation for edge devices and fine-tunes Llama-2-7B on a Jetson GPU~\cite{pockengine}. The open-source HexaTrain project trains small transformers on Hexagon NPUs~\cite{hexatrain}, and Fabric trains LoRA adapters on phone GPUs through Vulkan~\cite{qvac}. We run backpropagation through the frozen path of billion-parameter base models on the phone's inference NPU.

\paragraph{Rematerialization and offloading.}
Activation checkpointing recomputes intermediates to save memory~\cite{checkpoint}, and Checkmate and POET schedule rematerialization optimally or together with paging~\cite{checkmate,poet}; they complement our checkpoint budget. ZeRO-Offload divides training state and work between GPU and CPU~\cite{zerooffload}, work on AMD client NPUs offloads training's matrix products to the NPU~\cite{amd}, and Flying Serving switches an LLM server between data and tensor parallelism at run time as load changes~\cite{flyingserving}. Our partition also reuses resident inference weights and handles the engine's FP16 output range.

\paragraph{Quantization and gradient range.}
GPTQ and AWQ quantize weights~\cite{gptq,awq}, and LLM.int8() and SmoothQuant handle activation outliers~\cite{llmint8,smoothquant}, all for output quality; we set weight precision per projection by gradient sensitivity. Mixed-precision training keeps FP32 master state and scales the loss~\cite{mixedprecision}; our range guard acts per row and retries partial-result overflow.

\section{Experimental Configuration}\label{app:setup}

This appendix details the models, workloads, and training settings summarized in Section~\ref{sec:setup}. Table~\ref{tab:setup} gives the device and operating conditions for each experiment; baseline-specific configurations are described in Section~\ref{sec:setup} and Appendix~\ref{app:baselines}.

\subsection{Models and architecture}

The model set covers hybrid Qwen3.5 backbones~\cite{qwen35}, which interleave GDN recurrence and attention, and attention-only Qwen3~\cite{qwen3} and Llama~\cite{llama32} backbones. Table~\ref{tab:model_setup} summarizes their roles. Qwen3.5-9B is used for the weight-residency check in Section~\ref{sec:cost}.

\begin{table}[htb]
\caption{Models and evaluation roles.}
\label{tab:model_setup}
\centering\tabstyle
\begin{tabular}{@{}lll@{}}
\toprule
Model & Architecture & Evaluation role\\
\midrule
Q3.5-4B & Hybrid & Nightly; system tests\\
Q3.5-2B & Hybrid & LaMP-2; scaling\\
Q3.5-0.8B & Hybrid & Scaling\\
Q3-4B & Attention & LaMP-2\\
L3.2-3B & Attention & LaMP-2; phone GPU\\
\bottomrule
\end{tabular}
\end{table}

Q3.5-4B has 32 layers: 24 GDN and eight attention layers. Its text backbone has hidden width 2,560, MLP intermediate width 9,216, 16 attention heads, four key/value heads, and a vocabulary of 248,320 tokens. The 248 frozen projections cost 3.569 billion multiply-adds per token per forward pass. Three full traversals for forward, activation recomputation, and input gradients therefore require about 21.4 GFLOP per token. Frozen-weight formats follow the gradient-sensitive policy in Section~\ref{sec:policy}; the smaller Qwen3.5 models execute their 16-output gates on the CPU with complete forward and backward.

\subsection{Workloads and data}

\noindent\textbf{Nightly personalization.}
The nightly workload continues the previous night's Q3.5-4B adapter on 882 examples labeled on the phone. Two epochs contain 1,483,048 training tokens and 222 optimizer steps. Evaluation uses a 175-item next-day set and a 510-item held-out set, as described in Appendix~\ref{app:quality}.

\noindent\textbf{LaMP-2.}
LaMP-2~\cite{lamp} is a movie-tag classification benchmark with 3,820 training records and 692 development records. Each example includes two profile examples retrieved by BM25. Q3.5-2B, Q3-4B, and L3.2-3B each train a fresh adapter for one epoch, giving 478 optimizer steps. Task-quality metrics and GPU-reference comparisons follow Section~\ref{sec:setup} and Appendix~\ref{app:quality}.

\subsection{Training settings}

Table~\ref{tab:training_setup} lists the shared first-order LoRA settings for the nightly and LaMP-2 workloads. The CPU training dtype is BF16; NPU weight formats and arithmetic follow Section~\ref{sec:backward}. Both workloads supervise only answer positions. Their epoch counts and adapter initialization differ as described above.

\begin{table}[htb]
\caption{Shared first-order LoRA training settings.}
\label{tab:training_setup}
\centering\tabstyle
\begin{tabular}{@{}ll@{}}
\toprule
Parameter & Setting\\
\midrule
LoRA rank & 16\\
LoRA $\alpha$ & 32\\
LoRA dropout & 0.05\\
Bias adaptation & None\\
Learning rate & $10^{-4}$\\
Learning-rate schedule & Cosine\\
Training dtype & BF16\\
Maximum sequence length & 2,048 tokens\\
Microbatch size & 1 record\\
Gradient accumulation & 8 microbatches per step\\
Supervision & Answer positions only\\
\bottomrule
\end{tabular}
\end{table}

For Q3-4B and L3.2-3B, LoRA targets seven projection types: \texttt{q\_proj}, \texttt{k\_proj}, \texttt{v\_proj}, \texttt{o\_proj}, \texttt{gate\_proj}, \texttt{up\_proj}, and \texttt{down\_proj}. The Q3.5-4B nightly and Q3.5-2B LaMP-2 adapters also target \texttt{in\_proj\_qkv}, \texttt{in\_proj\_z}, \texttt{in\_proj\_a}, \texttt{in\_proj\_b}, and \texttt{out\_proj} in GDN layers. GPU-trained quality references use the same training recipe, with four dropout seeds for the nightly update and two for LaMP-2; Section~\ref{sec:setup} describes the seed comparison and baseline differences.

\end{document}